\documentclass[nohyperref]{article}

\usepackage{microtype}
\usepackage{graphicx}
\usepackage{booktabs} 

\usepackage{hyperref}

\PassOptionsToPackage{dvipsnames}{xcolor}
\PassOptionsToPackage{font=small}{caption}

\usepackage[accepted]{icml2022}
\makeatletter
\renewcommand{\printAffiliationsAndNotice}[1]{%
\stepcounter{@affiliationcounter}%
{\let\thefootnote\relax\footnotetext{\hspace*{-\footnotesep}\ifdefined\isaccepted #1\fi%
\forloop{@affilnum}{1}{\value{@affilnum} < \value{@affiliationcounter}}{
\textsuperscript{\arabic{@affilnum}}\ifcsname @affilname\the@affilnum\endcsname%
\csname @affilname\the@affilnum\endcsname%
\else
{\bf AUTHORERR: Missing \textbackslash{}icmlaffiliation.}
\fi
}.
\ifdefined\icmlcorrespondingauthor@text
Correspondence to: \icmlcorrespondingauthor@text. Work done while Tongzhou Wang was an intern at Meta AI.
\else
{\bf AUTHORERR: Missing \textbackslash{}icmlcorrespondingauthor.}
\fi

\ \\
\Notice@String
}
}
}
\makeatother

\usepackage{amsmath}
\usepackage{amssymb}
\usepackage{mathtools}
\usepackage{amsthm}
\usepackage[shortlabels]{enumitem}
\usepackage{subcaption}
\usepackage{afterpage}
\usepackage{makecell}
\usepackage{booktabs} 

\usepackage[capitalize,noabbrev]{cleveref}
\PassOptionsToPackage{numbers}{natbib}
\usepackage{macros}
\ExplSyntaxOn
\cs_new_eq:NN \fpeval \fp_eval:n
\ExplSyntaxOff
\newcommand {\robodesk}{\texorpdfstring{\texttt{RoboDesk}\@\xspace}{RoboDesk}}

\definecolor{myorange}{RGB}{230,97,0}
\newcommand{\new}[1]{{\color{red}#1}}
\renewcommand{\new}[1]{{#1}}

\newcommand{\hide}[1]{}

\newcommand{\exul}[1]{\expandafter\ul\expandafter{#1}}
\newcommand{\ctrl}{controllable\xspace}
\newcommand{\Ctrl}{Controllable\xspace}
\newcommand{\rewrel}{reward-relevant\xspace}

\newcommand{\nctrl}{uncontrollable\xspace}
\newcommand{\nCtrl}{Uncontrollable\xspace}
\newcommand{\nrewrel}{reward-irrelevant\xspace}

\setlist[itemize]{leftmargin=11pt,topsep=-3pt,itemsep=-3pt,parsep=3pt}
\setlist[itemize]{leftmargin=11pt,topsep=-2.5pt,itemsep=-0.5pt,parsep=3pt}

\usepackage{etoolbox}
\newcommand{\zerodisplayskips}{%
  \setlength{\abovedisplayskip}{3pt}%
  \setlength{\belowdisplayskip}{4pt}%
  \setlength{\abovedisplayshortskip}{3pt}%
  \setlength{\belowdisplayshortskip}{4pt}%
 }
\renewcommand{\zerodisplayskips}{%
  \setlength{\abovedisplayskip}{6pt}%
  \setlength{\belowdisplayskip}{8pt}%
  \setlength{\abovedisplayshortskip}{6pt}%
  \setlength{\belowdisplayshortskip}{8pt}%
 }
\appto{\normalsize}{\zerodisplayskips}
\appto{\small}{\zerodisplayskips}
\appto{\footnotesize}{\zerodisplayskips}

\usepackage{todonotes}

\icmltitlerunning{Denoised MDPs}

\title{\resizebox{\linewidth}{!}{\LARGE \bf Denoised MDPs: Learning World Models Better Than the World Itself}\vspace{-0.24cm}}

\author{
  \makebox[\textwidth][c]{%
  \resizebox{1.05\linewidth}{!}{%
  \begin{tabular}{c}
  \begin{tabular}{cccccc}
  \bf Tongzhou Wang$^{1}$  & \bf Simon S. Du$^2$ & \bf Antonio Torralba$^1$ & \bf Phillip Isola$^{1}$  & \bf Amy Zhang$^{34}$ & \bf Yuandong Tian$^4$
  \end{tabular}\\[0.8ex]
  \begin{tabular}{cccc}
  $^{1}$MIT CSAIL & $^{2}$University of Washington & $^{3}$UC Berkeley & $^{4}$Meta AI
  \end{tabular}
  \end{tabular}
  }%
  }
}

\date{}
 
\begin{document}

\twocolumn[
\icmltitle{Denoised MDPs: Learning World Models Better Than the World Itself}



\icmlsetsymbol{equal}{*}

\begin{icmlauthorlist}
\icmlauthor{Tongzhou Wang}{mit}
\icmlauthor{Simon S. Du}{uw}
\icmlauthor{Antonio Torralba}{mit}
\icmlauthor{Phillip Isola}{mit}
\icmlauthor{Amy Zhang}{berkeley,metaai}
\icmlauthor{Yuandong Tian}{metaai}
\end{icmlauthorlist}

\icmlaffiliation{mit}{MIT CSAIL}
\icmlaffiliation{uw}{University of Washington}
\icmlaffiliation{berkeley}{UC Berkeley}
\icmlaffiliation{metaai}{Meta AI}

\icmlcorrespondingauthor{Tongzhou Wang}{tongzhou@mit.edu}

\icmlkeywords{Machine Learning, ICML}

\vskip 0.3in
]








\printAffiliationsAndNotice{}  

\begin{abstract}


The ability to separate signal from noise, and reason with clean abstractions, is critical to intelligence. With this ability, humans can efficiently perform real world tasks without considering all possible nuisance factors.
How can artificial agents do the same?  What kind of information can agents safely discard as noises?  In this work, we categorize information out in the wild into four types based on controllability and relation with reward, and formulate useful information as that which is both \emph{\ctrl} and \emph{\rewrel}. This framework clarifies the kinds information removed by various prior work on representation learning in reinforcement learning (RL), and leads to our proposed approach of learning a \emph{Denoised MDP} that explicitly factors out certain noise distractors. Extensive experiments on variants of DeepMind Control Suite and \robodesk demonstrate superior performance of our denoised world model over using raw observations alone, and over prior works, across  policy optimization control tasks as well as the non-control task of joint position regression.

\begingroup%
\fontsize{8.75pt}{10pt}\selectfont%
\vspace{-0.75pt}\noindent\begin{tabular}{@{}lr@{}}
\textbf{Project Page:}\hspace{-3.2em} & \hspace{-1em}\href{https://ssnl.github.io/denoised_mdp/}{\texttt{ssnl.github.io/denoised\_mdp}} \\[0.3ex]
\textbf{Code:} & \hspace{-1em}\href{https://github.com/facebookresearch/denoised_mdp/}{\texttt{github.com/facebookresearch/denoised\_mdp}} \\
\end{tabular}%
\endgroup%
\vspace*{-10.5pt}
\end{abstract}

\section{Introduction}\label{sec:introduction}

The real world provides us a plethora of information, from microscopic physical interactions to abstracted semantic signals such as the latest {COVID-19} news. Fortunately, processing each and every signal is unnecessary (and also impossible). In fact, any particular reasoning or decision often only relies on a small portion of information.

\textit{Imagine waking up and wanting to embrace some sunlight.
As you open the curtain, a nearby resting bird is scared away and you are pleasantly met with a beautiful sunny day. Far away, a jet plane is slowly flying across the sky.}

\begin{figure}
\centering
\includegraphics[trim=52 21 27 10, scale=0.545]{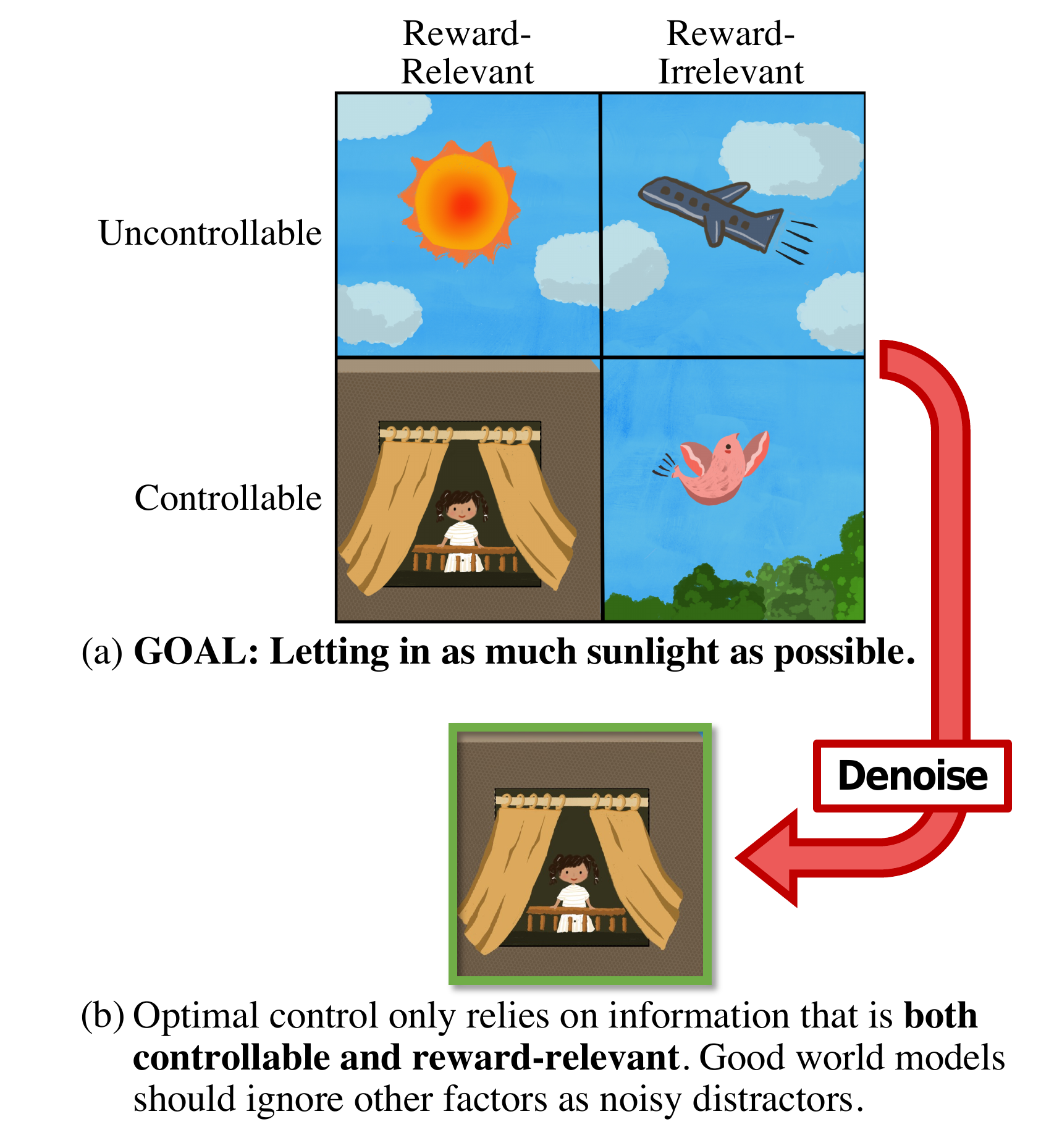}
\caption{\textbf{Illustrative example: (a)} Four distinct kinds of information in the scenario described in \Cref{sec:introduction}, where the person desires to increase the amount of sunlight let into the room. Their opening of the curtain scares away the bird. \textbf{(b)} A denoised world model only includes a small subset of all information.}\label{fig:intro-sunlight-2x2-example}%
\vspace{-5pt}
\end{figure}

This may seem a simple activity, but in fact highlights four distinct types of information (see \Cref{fig:intro-sunlight-2x2-example}), with respect to the goal of letting in as much sunlight as possible: \begin{itemize}[topsep=-4pt,itemsep=-0.5pt,parsep=1.5pt]
    \item {\textbf{\Ctrl and \rewrel}:}
    curtain, influenced by actions and affecting incoming sunlight;
    \item {\textbf{\Ctrl and \nrewrel}:}
    bird, influenced by actions but not affecting sunlight;
    \item {\textbf{\nCtrl and \rewrel}:}
    weather, independent with actions but affecting \hide{the amount of }sunlight;
    \item {\textbf{\nCtrl and \nrewrel}:}
    plane, independent with both actions and the sunlight.
\end{itemize}


Our optimal actions towards the goal, however, only in fact depend on information that is {\textbf{\ctrl and \rewrel}}, and the three other kinds of information are merely \emph{noise distractors}. Indeed, no matter how much natural sunlight there is outside, or how the plane and the bird move, the best plan is always to open up the curtain.

When performing a particular task, we humans barely think about the other three types of information, and usually only plan on how our actions affect information that is {\textbf{\ctrl and \rewrel}}. Our mental model is an abstract and condensed version of the real world that is actually \emph{better} suited for the task.

The notion of better model/data is ubiquitous in data science and machine learning. Algorithms rarely perform well on raw noisy real data. The common approach is to perform data cleaning and feature engineering, where we manually select the useful signals based on prior knowledge and/or heuristics. Years of research have identified ways to extract good features for computer vision \citep{lowe1999object,donahue2014decaf}, natural language processing \citep{elman1990finding,mikolov2013efficient}, reinforcement learning (RL) \citep{mahadevan2007proto,bellemare2019geometric}, \etc. Similarly, system identification aligns real observation with a predefined set of abstract signals/states. Yet for tasks in the wild (in the general form of (partially observable) Markov Decision Processes), there can be very little prior knowledge of the optimal set of signals.
In this work, we ask: can we infer and extract these signals automatically, in the form of a learned world model?

The general idea of a mental world model have long been under active research in philosophy and social science \citep{craik1952nature,dennett1975law}, cognitive science, where an intuitive physics model is hypothesized to be core in our planning capabilities \citep{spelke2007core}, and in reinforcement learning, where various methods investigate state abstractions for faster and better learning \citep{sutton1991dyna,sutton1981adaptive}.

In this work, we explore this idea within the context of machine learning and reinforcement learning, where we aim to make concrete the different types of information in the wild, and automatically learn a world model that removes noise distractors and is beneficial for both control (\ie, policy optimization) and non-control tasks. Toward this goal, our contributions are \begin{itemize}
    \item We categorize information into four distinct kinds as in \Cref{fig:intro-sunlight-2x2-example}, and review prior approaches under this framework (\Cref{sec:information-world}).
    \item Based on the above framework, we propose Denoised MDPs, a method for learning world models with certain distractors removed (\Cref{sec:denoise}).
    \item Through experiments in DeepMind Control Suite and \robodesk environments, we demonstrate superior performance of policies learned our method, across many distinct types of noise distractors (\Cref{sec:dmc,sec:robodesk}).
    \item We show that Denoised MDP is also beneficial beyond control objectives, improving the supervised task of robot joint position regression (\Cref{sec:robodesk}).
\end{itemize}


\section{Different Types of Information in the Wild}\label{sec:information-world}

\begin{figure*}[ht]
    \vspace{5pt}
    \centering
    \begin{subfigure}[t]{4.9cm}
        \centering
        \scalebox{0.9}{
        \hspace{-0.1cm}\begin{tikzpicture}[bayes_net, node distance = 0.5cm, every node/.style={inner sep=0, minimum size = 0.9cm}]
            \node[main_node] (s) {$s$};
            \node[draw=none] (dummy) [below=-0.82cm of s] {};
            \node[main_node] (sp) [right = 0.4cm of s] {$s'$};
            \node[main_node, draw=none, text opacity=0] (aheightholder) [above right = 3.85cm and 0.1cm of s] {$a$};
            \node[main_node] (a) [above right = 1.15cm and -0.175cm of s] {$a$};
            \node[main_node] (r) [above right = 1.15cm and -0.4cm of sp] {$r$};

            \node[minimum size = 1.075cm, draw, inner sep=0, label={[scale=1.2]center:$s$}] (cbr) [above right = -0.75cm and 1.1cm  of sp] {};
            \node[minimum size = 1.075cm, draw, label={[scale=1.2]center:$s$}] (cbrb) [right = -0.03cm of cbr] {};
            \node[minimum size = 1.075cm, draw, label={[scale=1.2]center:$s$}] (cr) [above = -0.03cm of cbr] {};
            \node[minimum size = 1.075cm, draw, label={[scale=1.2]center:$s$}] (crb) [right = -0.03cm of cr] {};
            \node[draw=none, minimum width=.7cm, label={[scale=0.9]center:\textbf{Ctrl}}] () [left = 0cm of cr] {};
            \node[draw=none, minimum width=.7cm, label={[scale=0.9]center:\textbf{\textoverline{Ctrl}}}] () [left = 0cm of cbr] {};
            \node[draw=none, minimum height=.33cm, label={[scale=0.9]center:\textbf{Rew}}] () [above = 0cm of cr] {};
            \node[draw=none, minimum height=.4cm, label={[scale=0.9]center:\textbf{\textoverline{Rew}}}] () [above = 0cm of crb] {};

            \tikzset{myptr/.style={decoration={markings,mark=at position 1 with %
                {\arrow[scale=0.825,>=stealth']{>}}},postaction={decorate}}}
            \path[]
            (s) edge[-, myptr, line width=1.2pt] [right] node [left=3pt] {} (sp)
            (a) edge[-, myptr, line width=1.2pt] [right] node [below=3pt] {} (sp)
            (sp) edge[-, myptr, line width=1.2pt] [right] node [left=3pt] {} (r)
            ;
        \end{tikzpicture}
        }\vspace{-5pt}%
        \caption{Transition without useful structure. $s$ may contain any type of information.}\label{fig:mdp-grid-regular}
    \end{subfigure}%
    \hfill%
    \begin{subfigure}[t]{5.6cm}
        \centering
        \scalebox{0.9}{
        \hspace{-0.1cm}\begin{tikzpicture}[bayes_net, node distance = 0.5cm, every node/.style={inner sep=0, minimum size = 0.9cm}]
            \node[main_node] (s) {$s$};
            \node[main_node, color=MyColorYRB] (y) [above = 0.45cm of s] {$y_{\overbar{R}}$};
            \node[main_node, color=MyColorYR] (yR) [above = 0.225cm of y] {$y_R$};
            \node[main_node, color=MyColorX] (x) [above = 0.225cm of yR] {$x$};
            \node[main_node] (sp) [right = 0.8cm of s] {$s'$};
            \node[main_node, color=MyColorX] (xp) [right = 0.8cm of x] {$x'$};
            \node[main_node, color=MyColorYR] (yRp) [right = 0.8cm of yR] {$y_R'$};
            \node[main_node, color=MyColorX] (rx) [right = 0.385cm of xp] {$r_x$};
            \node[main_node, color=MyColorYR] (ry) [right = 0.385cm of yRp] {$r_y$};
            \node[main_node, color=MyColorYRB] (yp) [right = 0.8cm of y] {$y'_{\overbar{R}}$};
            \node[main_node] (a) [above right = 4cm and 0.1cm of s] {$a$};
            \node[minimum width=0.335cm, minimum height=0.75cm, draw] (plus) [above right = -0.15cm and 0.35cm of ry] {$+$};
            \node[main_node] (r) [right = 0.25cm of plus] {$r$};


            \tikzset{myptr/.style={decoration={markings,mark=at position 1 with %
                {\arrow[scale=0.825,>=stealth']{>}}},postaction={decorate}}}
            \path[]
            (x) edge[-, myptr, line width=1.2pt] [right] node [left=3pt] {} (xp)
            (xp) edge[-, myptr, line width=1.2pt] [right] node [left=3pt] {} (rx)
            (rx.345) edge[-, myptr, line width=1.2pt] [right] node [left=3pt] {} (plus)
            (a) edge[-, myptr, line width=1.2pt] [right] node [below=3pt] {} (xp)
            (y) edge[-, myptr, line width=1.2pt] [right] node [left=3pt] {} (yp)
            (yR) edge[-, myptr, line width=1.2pt] [right] node [left=3pt] {} (yRp)
            (yRp) edge[-, myptr, line width=1.2pt] [right] node [left=3pt] {} (ry)
            (ry.15) edge[-, myptr, line width=1.2pt] [right] node [left=3pt] {} (plus)
            (plus) edge[-, myptr, line width=1.2pt] [right] node [left=3pt] {} (r)
            (y) edge[-, myptr, line width=1.2pt, shorten <= 0.13cm] [right] node [midway, above=3pt] {} (s)
            (yp) edge[-, myptr, line width=1.2pt, shorten <= 0.13cm] [right] node [midway, above=3pt] {} (sp);


            \node[minimum size = 1.075cm, draw] (cbr) [above right = -0.75cm and 1.3cm  of sp] {};
            \node[minimum size = 1.075cm, draw] (cbrb) [right = -0.03cm of cbr] {};
            \node[minimum size = 1.075cm, draw] (cr) [above = -0.03cm of cbr] {};
            \node[minimum size = 1.075cm, draw] (crb) [right = -0.03cm of cr] {};

            \node[] (cbrx) [above = -0.875cm of cbr] {\cx};
            \node[] (crx) [above = -0.875cm of cr] {\cx};
            \node[] (crbx) [above = -0.875cm of crb] {\cx};
            \node[] (cbrbx) [above = -0.875cm of cbrb] {\cx};

            \node[text opacity=0] (cbry) at ([shift={(-0.215,-0.335)}]cbrx) {\cyrb};
            \node[text opacity=0] (cry) at ([shift={(-0.215,-0.335)}]crx) {\cyrb};
            \node[text opacity=0] (crby) at ([shift={(-0.215,-0.335)}]crbx) {\cyrb};
            \node[] (cbrby) at ([shift={(-0.215,-0.335)}]cbrbx) {\cyrb};

            \node[] (cbryr) at ([shift={(0.275,-0.319)}]cbrx) {\cyr};
            \node[text opacity=0] (cryr) at ([shift={(0.275,-0.319)}]crx) {\cyr};
            \node[text opacity=0] (crbyr) at ([shift={(0.275,-0.319)}]crbx) {\cyr};
            \node[] (cbrbyr) at ([shift={(0.275,-0.319)}]cbrbx) {\cyr};

            \node[draw=none, minimum width=.7cm, label={[scale=0.9]center:\textbf{Ctrl}}] () [left = 0cm of cr] {};
            \node[draw=none, minimum width=.7cm, label={[scale=0.9]center:\textbf{\textoverline{Ctrl}}}] () [left = 0cm of cbr] {};
            \node[draw=none, minimum height=.33cm, label={[scale=0.9]center:\textbf{Rew}}] () [above = 0cm of cr] {};
            \node[draw=none, minimum height=.4cm, label={[scale=0.9]center:\textbf{\textoverline{Rew}}}] () [above = 0cm of crb] {};

            \begin{scope}[on background layer]
                \draw[fill=black, fill opacity=0.08, color=black, -, dashed, shorten >= 0pt] \convexpath{x,y}{0.575cm};
                \draw[fill=black, fill opacity=0.08, color=black, -, dashed, shorten >= 0pt] \convexpath{xp,yp}{0.575cm};
            \end{scope}
        \end{tikzpicture}%
        }\vspace{-5pt}%
        \caption{Transition that factorizes out \nctrl information in {\cyrb} and \cyr. 
        }%
        \label{fig:mdp-grid-xy}
    \end{subfigure}%
    \hfill%
    \begin{subfigure}[t]{5.6cm}
        \centering
        \scalebox{0.9}{
        \hspace{-0.1cm}\begin{tikzpicture}[bayes_net, node distance = 0.5cm, every node/.style={inner sep=0, minimum size = 0.9cm}]
            \node[main_node] (s) {$s$};
            \node[main_node, color=MyColorZ] (z) [above = 0.45cm of s] {$z$};
            \node[main_node, color=MyColorY] (y) [above = 0.225cm of z] {$y$};
            \node[main_node, color=MyColorX] (x) [above = 0.225cm of y] {$x$};
            \node[main_node] (sp) [right = 0.8cm of s] {$s'$};
            \node[main_node, color=MyColorX] (xp) [right = 0.8cm of x] {$x'$};
            \node[main_node, color=MyColorY] (yp) [right = 0.8cm of y] {$y'$};
            \node[main_node, color=MyColorX] (rx) [right = 0.385cm of xp] {$r_x$};
            \node[main_node, color=MyColorY] (ry) [right = 0.385cm of yp] {$r_y$};
            \node[main_node, color=MyColorZ] (zp) [right = 0.8cm of z] {$z'$};
            \node[main_node] (a) [above right = 4cm and 0.1cm of s] {$a$};
            \node[minimum width=0.335cm, minimum height=0.75cm, draw] (plus) [above right = -0.15cm and 0.35cm of ry] {$+$};
            \node[main_node] (r) [right = 0.25cm of plus] {$r$};

            \tikzset{myptr/.style={decoration={markings,mark=at position 0.99999 with %
                {\arrow[scale=0.825,>=stealth']{>}}},postaction={decorate}}}
            \path[]
            (x) edge[-, myptr, line width=1.2pt] [right] node [left=3pt] {} (xp)
            (xp) edge[-, myptr, line width=1.2pt] [right] node [left=3pt] {} (rx)
            (rx.345) edge[-, myptr, line width=1.2pt] [right] node [left=3pt] {} (plus)
            (a) edge[-, myptr, line width=1.2pt] [right] node [below=3pt] {} (xp)
            (y) edge[-, myptr, line width=1.2pt] [right] node [left=3pt] {} (yp)
            (yp) edge[-, myptr, line width=1.2pt] [right] node [left=3pt] {} (ry)
            (ry.15) edge[-, myptr, line width=1.2pt] [right] node [left=3pt] {} (plus)
            (plus) edge[-, myptr, line width=1.2pt] [right] node [left=3pt] {} (r)
            (a) edge[-, myptr, line width=1.2pt, bend right=20] [right] node [midway, above=3pt] {} (zp)
            (z) edge[-, myptr, line width=1.2pt] [right] node [midway, above=3pt] {} (zp)
            (yp) edge[-, myptr, line width=1.2pt] [right] node [midway, above=3pt] {} (zp)
            (z) edge[-, myptr, line width=1.2pt, shorten <= 0.13cm] [right] node [midway, above=3pt] {} (s)
            (xp) edge[-, myptr, line width=1.2pt, bend left=35] [left] node [midway, above=3pt] {} (zp)
            (zp) edge[-, myptr, line width=1.2pt, shorten <= 0.13cm] [right] node [midway, above=3pt] {} (sp);


            \node[minimum size = 1.075cm, draw] (cbr) [above right = -0.75cm and 1.3cm  of sp] {};
            \node[minimum size = 1.075cm, draw] (cbrb) [right = -0.03cm of cbr] {};
            \node[minimum size = 1.075cm, draw] (cr) [above = -0.03cm of cbr] {};
            \node[minimum size = 1.075cm, draw] (crb) [right = -0.03cm of cr] {};

            \node[] (cbrx) [above = -0.875cm of cbr] {\cx};
            \node[] (crx) [above = -0.875cm of cr] {\cx};
            \node[] (crbx) [above = -0.875cm of crb] {\cx};
            \node[] (cbrbx) [above = -0.875cm of cbrb] {\cx};

            \node[] (cbry) at ([shift={(-0.2,-0.33)}]cbrx) {\cy};
            \node[text opacity=0] (cry) at ([shift={(-0.2,-0.33)}]crx) {\cy};
            \node[text opacity=0] (crby) at ([shift={(-0.2,-0.33)}]crbx) {\cy};
            \node[] (cbrby) at ([shift={(-0.2,-0.33)}]cbrbx) {\cy};

            \node[text opacity=0] (cbrz) at ([shift={(0.215,-0.3075)}]cbrx) {\cz};
            \node[text opacity=0] (crz) at ([shift={(0.215,-0.3075)}]crx) {\cz};
            \node[] (crbz) at ([shift={(0.215,-0.3075)}]crbx) {\cz};
            \node[] (cbrbz) at ([shift={(0.215,-0.3075)}]cbrbx) {\cz};

            \node[draw=none, minimum width=.7cm, label={[scale=0.9]center:\textbf{Ctrl}}] () [left = 0cm of cr] {};
            \node[draw=none, minimum width=.7cm, label={[scale=0.9]center:\textbf{\textoverline{Ctrl}}}] () [left = 0cm of cbr] {};
            \node[draw=none, minimum height=.33cm, label={[scale=0.9]center:\textbf{Rew}}] () [above = 0cm of cr] {};
            \node[draw=none, minimum height=.4cm, label={[scale=0.9]center:\textbf{\textoverline{Rew}}}] () [above = 0cm of crb] {};

            \begin{scope}[on background layer]
                \draw[fill=black, fill opacity=0.08, color=black, -, dashed, shorten >= 0pt] \convexpath{x,z}{0.575cm};
                \draw[fill=black, fill opacity=0.08, color=black, -, dashed, shorten >= 0pt] \convexpath{xp,zp}{0.575cm};
            \end{scope}
        \end{tikzpicture}
        }\vspace{-5pt}%
        \caption{Transition that factorizes out \nctrl $\cy$ and \nrewrel $\cz$.}%
        \label{fig:mdp-grid-xyz}
    \end{subfigure}\vspace{-6.5pt}%
    \caption{MDP transition structures consisting of dynamics and reward functions. Unlike the regular structure of \textbf{(a)}, \textbf{(b,~c)} factorized (yet still general) structures  inherently separate information into \ctrl (\textbf{Ctrl}) versus \nctrl (\textbf{\textoverline{Ctrl}}), and \rewrel (\textbf{Rew}) versus \nrewrel (\textbf{\textoverline{Rew}}). Presence of a variable in a cell means \emph{possible} containing of respective information. \Eg, in \textbf{(c)}, $\cz$ can only contain \nrewrel information\hide{, regardless of controllability}. In \textbf{(b,~c)}, the $\cx$ dynamics form an MDP with less noise and sufficient for optimal planning. Our Denoised MDP (see \Cref{sec:denoise}) is based on these two factorizations.}\label{fig:mdp-transition-grid}
    \vspace{-4.5pt}
\end{figure*}%

In \Cref{sec:introduction}, we illustrated the four types of information available in the wild \wrt a task. Here we make these notions more concrete, and relate them to existing works.

For generality, we consider tasks in the form of Markov Decision Processes (MDPs), described in the usual manner: $\mathcal{M} \trieq (\mathcal{S}, \mathcal{A}, R, P, p_{s_0})$ \citep{puterman1994mdp}, where $\mathcal{S}$ is the state space, $\mathcal{A}$ is the action space, $R \colon \mathcal{S} \rightarrow \Delta([0, r_\mathsf{max}])$ defines the reward random variable $R(s')$ received for arriving at state $s' \in \mathcal{S}$, $P \colon \mathcal{S} \times \mathcal{A} \rightarrow \Delta(\mathcal{S})$ is the transition dynamics, 
and $p_{s_0} \in \Delta(\mathcal{S})$ defines the distribution of initial state.  We use $\Delta(A)$ to denote the set of all distributions over $A$.
$P$ and $R$ define the most important components of a MDP: the transition dynamics $\mathbb{P}[s' \given s,a]$ and the reward function $\mathbb{P}[r \given s']$. Usually, the objective is to find a policy $\pi \colon \mathcal{S} \rightarrow \Delta(\mathcal{A})$ acting based on current state, that maximizes the expected cumulative (discounted) reward.

Indeed, MDPs provide a  general formulation that encompasses many tasks.
In fact, the entire real world may be viewed as an MDP with a rich state/observation space $\mathcal{S}$ that contains all possible information/signal.
For an artificial agent to successfully perform real world tasks, it must be able to process observations that are incredibly rich and high-dimensional, such as visual or audio signals.

We characterize different types of information in such observations by considering  two intuitive notions of ``noisy and irrelevant'' signals: (1) \nctrl information and (2) \nrewrel information. Such factors can often be ignored without affecting optimal control, and are referred to as \emph{noise distractors}.

To understand their roles in MDPs, we study different formulations of the transition dynamics and reward functions, and show how different structures naturally leads to decompositions that may help identify such distractors. Removing these distractors can thus \emph{transform the original noisy MDP to a clean denoised one}, to be used in downstream tasks.



For starters, the most generic transition model in \Cref{fig:mdp-grid-regular} has little to no structure. The state $s$ can contain both the useful signals and noise distractors. Therefore, it is not directly useful for extracting important information.

\subsection{Controllability}

Intuitively, if something is not controllable, an agent might be able to do well without considering it. Yet it is not enough to only require some variable to be unaffected by actions (\eg, wind directions should not be ignored while sailing). Instead, we focus on
factors that simply evolve on their own, without influencing or being influenced by others.


Not all such information can be safely ignored, as they still may affect reward (\eg, traffic lights when driving). Fortunately, in the usual objective of maximizing expected  return, we can ignore ones that only additively affect reward.

Concretely, if an MDP transition can be represented in the form of  \Cref{fig:mdp-grid-xy}, we say variables $\cyrb$ and  $\cyr$ are \emph{\nctrl} information, as they evolve independently of actions and do not affect \emph{\ctrl} $\cx$. Here $\cyr$ (additively) affects reward, but can be ignored. One can safely discard both $\cyrb$ and  $\cyr$ as \emph{noise distractors}. Operating with the compressed MDP of only $\cx$ is sufficient for optimal control.

\subsection{Reward-Relevance}

Among \ctrl information, there can still be some that is completely unrelated to reward. In \Cref{fig:intro-sunlight-2x2-example}, the bird is affected by the opening curtain, but is irrelevant to the task of letting in sunlight. In such cases, the information can be safely discarded, as it does not affect the objective.

If an MDP transition can be represented in the form of  \Cref{fig:mdp-grid-xyz}, we say $\cz$ is \emph{\nrewrel} because it evolves by potentially using everything (\ie, all latent variables and actions), but crucially \emph{does not affect anything but itself}.


Similar to \nctrl information, $\cz$ (and $\cy$) is a \emph{noise distractor} that can be discarded. The compressed MDP of only $\cx$ contains all signals needed for optimal control.


\subsection{Which Information Do Existing Methods Learn?}\label{sec:grid-existing}

\begin{figure}[t]
\centering
\vspace{-5.5pt}
\hspace*{-0.12cm}\scalebox{0.88}{
\begin{tikzpicture}[bayes_net, node distance = 0.5cm, every node/.style={inner sep=0, minimum size = 0.9cm}]
    \node[] (anchor)[] {};

    \node[below = -1.2cm of anchor] (recon_text) [text opacity=1, align=center, minimum width=4.45cm, minimum height=2cm]
    {\small\bf%
    \shortstack{Reconstruction-Based\\Model-Based RL\\\rm (\eg, SLAC \citep{lee2019stochastic},\\\rm Dreamer \citep{hafner2019dream})}
    };

    \node[right = -0.15cm of recon_text] (recon_desc) [text opacity=1, align=center, text width=2.5cm, minimum height=2cm, execute at begin node=\setlength{\baselineskip}{2ex}]
    {\small\bf
    \color{Magenta}\ul{Model-Based}
    };

    \begin{scope}[shift={($(recon_text.east)+(3.5cm,-0.4cm)$)}]
        \node[minimum size = 0.63cm, draw, fill=ForestGreen, fill opacity=0.5, text opacity=1] (cbr) [] {\cmark};
        \node[minimum size = 0.63cm, draw, fill=ForestGreen, fill opacity=0.5, text opacity=1] (cbrb) [right = -0.03cm of cbr] {\cmark};
        \node[minimum size = 0.63cm, draw, fill=ForestGreen, fill opacity=0.5, text opacity=1] (cr) [above = -0.03cm of cbr] {\cmark};
        \node[minimum size = 0.63cm, draw, fill=ForestGreen, fill opacity=0.5, text opacity=1] (crb) [right = -0.03cm of cr] {\cmark};

        \node[draw=none, minimum width=.7cm, label={[scale=0.9]center:\textbf{Ctrl}}] () [left = 0cm of cr] {};
        \node[draw=none, minimum width=.7cm, label={[scale=0.9]center:\textbf{\textoverline{Ctrl}}}] () [left = 0cm of cbr] {};
        \node[draw=none, minimum height=.33cm, label={[scale=0.9]center:\textbf{Rew}}] () [above = 0cm of cr] {};
        \node[draw=none, minimum height=.4cm, label={[scale=0.9]center:\textbf{\textoverline{Rew}}}] () [above = 0cm of crb] {};
    \end{scope}

    \node[below = -0.2cm  of  recon_text] (bisim_text) [text opacity=1, align=center, minimum width=4.45cm, minimum height=2cm]
    {\small\bf%
    \shortstack{Bisimulation\\\rm (\eg, \citet{ferns2004metrics},\\\rm\citet{castro2020scalable}, \citet{zhang2020learning})}
    };

    \node[right = -0.15cm of bisim_text] (bisim_desc) [text opacity=1, align=center, text width=2.5cm, minimum height=2cm, execute at begin node=\setlength{\baselineskip}{2ex}]
    {\small\bf
    \color{RoyalPurple}\ul{Model-Free}
    };

    \begin{scope}[shift={($(bisim_text.east)+(3.5cm,-0.4cm)$)}]
        \node[minimum size = 0.63cm, draw, fill=ForestGreen, fill opacity=0.5, text opacity=1] (cbr) [] {\cmark};
        \node[minimum size = 0.63cm, draw, fill=BrickRed, fill opacity=0.5, text opacity=1] (cbrb) [right = -0.03cm of cbr] {\xmark};
        \node[minimum size = 0.63cm, draw, fill=ForestGreen, fill opacity=0.5, text opacity=1] (cr) [above = -0.03cm of cbr] {\cmark};
        \node[minimum size = 0.63cm, draw, fill=BrickRed, fill opacity=0.5, text opacity=1] (crb) [right = -0.03cm of cr] {\xmark};

        \node[draw=none, minimum width=.7cm, label={[scale=0.9]center:\textbf{Ctrl}}] () [left = 0cm of cr] {};
        \node[draw=none, minimum width=.7cm, label={[scale=0.9]center:\textbf{\textoverline{Ctrl}}}] () [left = 0cm of cbr] {};
        \node[draw=none, minimum height=.33cm, label={[scale=0.9]center:\textbf{Rew}}] () [above = 0cm of cr] {};
        \node[draw=none, minimum height=.4cm, label={[scale=0.9]center:\textbf{\textoverline{Rew}}}] () [above = 0cm of crb] {};
    \end{scope}

    \node[below = -0.2cm  of  bisim_text] (tia_text) [text opacity=1, align=center, minimum width=4.45cm, minimum height=2cm]
    {\small\bf%
    \shortstack{Task Informed\\Abstractions (TIA)\\\rm \citep{fu2021learning}}
    };

    \node[right = -0.15cm of tia_text] (tia_desc) [text opacity=1, align=center, text width=2.5cm, minimum height=2cm, execute at begin node=\setlength{\baselineskip}{2ex}]
    {\small\bf
    \color{Magenta}\ul{Model-Based}
    };

    \begin{scope}[shift={($(tia_text.east)+(3.5cm,-0.4cm)$)}]
        \node[minimum size = 0.63cm, draw, fill=ForestGreen, fill opacity=0.5, text opacity=1] (cbr) [] {\cmark};
        \node[minimum size = 0.63cm, draw, fill=BurntOrange, fill opacity=0.5, text opacity=1] (cbrb) [right = -0.03cm of cbr] {\textbf{?}};
        \node[minimum size = 0.63cm, draw, fill=ForestGreen, fill opacity=0.5, text opacity=1] (cr) [above = -0.03cm of cbr] {\cmark};
        \node[minimum size = 0.63cm, draw, fill=BurntOrange, fill opacity=0.5, text opacity=1] (crb) [right = -0.03cm of cr] {\textbf{?}};

        \node[draw=none, minimum width=.7cm, label={[scale=0.9]center:\textbf{Ctrl}}] () [left = 0cm of cr] {};
        \node[draw=none, minimum width=.7cm, label={[scale=0.9]center:\textbf{\textoverline{Ctrl}}}] () [left = 0cm of cbr] {};
        \node[draw=none, minimum height=.33cm, label={[scale=0.9]center:\textbf{Rew}}] () [above = 0cm of cr] {};
        \node[draw=none, minimum height=.4cm, label={[scale=0.9]center:\textbf{\textoverline{Rew}}}] () [above = 0cm of crb] {};
    \end{scope}

    \node[below = -0.2cm  of  tia_text] (ours1_text) [text opacity=1, align=center, minimum width=4.45cm, minimum height=2cm]
    {\small\bf%
    \shortstack{Denoised MDP\\(\Cref{fig:mdp-grid-xy} variant)\\\rm(Our method from \Cref{sec:denoise})}
    };

    \node[right = -0.15cm of ours1_text] (ours_desc) [text opacity=1, align=center, text width=2.5cm, minimum height=2cm, execute at begin node=\setlength{\baselineskip}{2ex}]
    {\small\bf
    \color{Magenta}\ul{Model-Based}
    };

    \begin{scope}[shift={($(ours1_text.east)+(3.5cm,-0.4cm)$)}]
        \node[minimum size = 0.63cm, draw, fill=BrickRed, fill opacity=0.5, text opacity=1] (cbr) [] {\xmark};
        \node[minimum size = 0.63cm, draw, fill=BrickRed, fill opacity=0.5, text opacity=1] (cbrb) [right = -0.03cm of cbr] {\xmark};
        \node[minimum size = 0.63cm, draw, fill=ForestGreen, fill opacity=0.5, text opacity=1] (cr) [above = -0.03cm of cbr] {\cmark};
        \node[minimum size = 0.63cm, draw, fill=ForestGreen, fill opacity=0.5, text opacity=1] (crb) [right = -0.03cm of cr] {\cmark};

        \node[draw=none, minimum width=.7cm, label={[scale=0.9]center:\textbf{Ctrl}}] () [left = 0cm of cr] {};
        \node[draw=none, minimum width=.7cm, label={[scale=0.9]center:\textbf{\textoverline{Ctrl}}}] () [left = 0cm of cbr] {};
        \node[draw=none, minimum height=.33cm, label={[scale=0.9]center:\textbf{Rew}}] () [above = 0cm of cr] {};
        \node[draw=none, minimum height=.4cm, label={[scale=0.9]center:\textbf{\textoverline{Rew}}}] () [above = 0cm of crb] {};
    \end{scope}

    \node[below = -0.2cm  of  ours1_text] (ours2_text) [text opacity=1, align=center, minimum width=4.45cm, minimum height=2cm]
    {\small\bf%
    \shortstack{Denoised MDP\\(\Cref{fig:mdp-grid-xyz} variant)\\\rm(Our method from \Cref{sec:denoise})}
    };

    \node[right = -0.15cm of ours2_text] (ours_desc) [text opacity=1, align=center, text width=2.5cm, minimum height=2cm, execute at begin node=\setlength{\baselineskip}{2ex}]
    {\small\bf
    \color{Magenta}\ul{Model-Based}
    };

    \begin{scope}[shift={($(ours2_text.east)+(3.5cm,-0.4cm)$)}]
        \node[minimum size = 0.63cm, draw, fill=BrickRed, fill opacity=0.5, text opacity=1] (cbr) [] {\xmark};
        \node[minimum size = 0.63cm, draw, fill=BrickRed, fill opacity=0.5, text opacity=1] (cbrb) [right = -0.03cm of cbr] {\xmark};
        \node[minimum size = 0.63cm, draw, fill=ForestGreen, fill opacity=0.5, text opacity=1] (cr) [above = -0.03cm of cbr] {\cmark};
        \node[minimum size = 0.63cm, draw, fill=BrickRed, fill opacity=0.5, text opacity=1] (crb) [right = -0.03cm of cr] {\xmark};

        \node[draw=none, minimum width=.7cm, label={[scale=0.9]center:\textbf{Ctrl}}] () [left = 0cm of cr] {};
        \node[draw=none, minimum width=.7cm, label={[scale=0.9]center:\textbf{\textoverline{Ctrl}}}] () [left = 0cm of cbr] {};
        \node[draw=none, minimum height=.33cm, label={[scale=0.9]center:\textbf{Rew}}] () [above = 0cm of cr] {};
        \node[draw=none, minimum height=.4cm, label={[scale=0.9]center:\textbf{\textoverline{Rew}}}] () [above = 0cm of crb] {};
    \end{scope}

    \node[below right = -0.076cm and -4.6cm of ours2_text] (legend) [text opacity=1, align=left, minimum width=9.2cm, minimum height=0.5cm, draw,fit margins={left=3pt,right=2.5pt,bottom=1.3pt,top=1.3pt}]
    {\small
    \shortstack[l]{%
    {\bf Information Grid Legend:}%
    \hspace{0.55cm}%
    \raisebox{-3pt}{\protect\tikz{%
    \node[fill=ForestGreen, fill opacity=0.5, rectangle, inner sep=1pt, minimum width=3ex, minimum height=3ex, draw] {\cmark}%
    }}~~Kept%
    \hspace{1.25cm}%
    \raisebox{-3pt}{\protect\tikz{%
    \node[fill=BrickRed, fill opacity=0.5, rectangle, inner sep=1pt, minimum width=3ex, minimum height=3ex, draw] {\xmark}%
    }}~~Reduced%
    \\[0.9ex]%
    \phantom{\bf Information Grid Legend:}\hspace{0.55cm}%
    \raisebox{1pt}{\protect\tikz{%
    \node[fill=BurntOrange, fill opacity=0.5, rectangle, inner sep=1pt, minimum width=3ex, minimum height=3ex, draw] {\textbf{?}}%
    }}%
    ~~\shortstack[l]{Depending on how the information\hspace*{-2pt}\\[-0.6ex]is integrated in observations}%
    }%
    };
\end{tikzpicture}%
}\vspace{-6pt}
\caption{
Categorization of information learned and removed by various methods with distinct formulations. 
}\label{fig:grid-methods}%
\vspace{-6pt}
\end{figure}

In RL, many prior work have explored state abstractions in some form. Here we cast several representative ones under the framework described above, and show which kinds of information they learn to remove, summarized in \Cref{fig:grid-methods}, together with our proposed method (explained in \Cref{sec:denoise}). Below we discuss each prior work in detail.

\myparagraph{Reconstruction-Based Model-Based RL.} Many model-based RL methods learn via reconstruction from a single latent code, often as a result of a variational formulation \citep{hafner2019dream,hafner2019learning,lee2019stochastic}. The latent code must try to compress all information present in the observation, and necessarily contains all types of information.

\myparagraph{Bisimulation.} Bisimulation defines a state abstraction where states aggregated together must have the same expected return and transition dynamics up to the abstraction \citep{givan2003equivalence}, and is known to optimally ignore \nrewrel information \citep{ferns2004metrics}. While its continuous version, bisimilation metric, is gaining popularity, learning them is computationally difficult \citep{modi2020sample}. Even with many additional assumptions,  it is generally only possible to learn an on-policy variant that loses the above guarantee \citep{castro2020scalable,zhang2020learning}.

\myparagraph{Task Informed Abstractions (TIA).} TIA \citep{fu2021learning} extends Dreamer by modelling two independent latent MDPs, representing signal and noise. The noise latent is enforced to be independent with reward and reconstruct the observation as well as possible. Reconstructions from each latent are composed together using an inferred mask in pixel-space, to form the full reconstruction for the reconstruction loss. Because of its special structure, TIA can remove \emph{\nrewrel} noise distractors that are present via pixel-wise composing two images from \emph{independent} processes (\eg, agent moving on a noisy background), but not general ones (\eg, a shaky camera affecting both the agent and the noisy background).

\new{\myparagraph{Predictive Information, Data Augmentation, etc.}Another set of researches learn state representation that only contains information useful for predicting future states (\eg, CPC \citep{oord2018representation} and PI-SAC \citep{lee2020predictive}) or augmented views of the current state (\eg, CURL \citep{laskin2020curl}). These methods \emph{do not guarantee} removal of any of the three redundant piece of information identified above. Non-i.i.d.~noises (\eg, people moving in background) are predictive of future and may be kept by CPC and PI-SAC. The performance of augmentation-based methods can critically rely on specific types of augmentation used and relevance to the tasks. As we show in experiments (see \Cref{sec:expr}), indeed they struggle to handle certain noise types.} 

\subsection{Possible Extensions to Further Factorizations}\label{sec:factorization-extension}

The above framework is sufficient for characterizing most prior work and related tasks, and can also be readily extended with further factorized transition structures. \Eg, if an independent process confounds a signal process and a noise process, fitting the \Cref{fig:mdp-grid-xyz} structure must group all three processes into $\cx$ (to properly model the dependencies). However, a further factorization shows that only considering the signal and the confounding processes is theoretically sufficient for control.  We leave such extensions as future work.

\section{Denoised MDPs}\label{sec:denoise}

\Cref{fig:mdp-grid-xy,fig:mdp-grid-xyz} show two special MDP structures that automatically identify certain information that can be ignored, leaving $\cx$ as the useful information (which also forms an MDP). This suggests a na\"{i}ve approach: directly fitting such structures to collected trajectories, and then extract $\cx$.

However, the same MDP dynamics and rewards can be decomposed as \Cref{fig:mdp-grid-xy,fig:mdp-grid-xyz} in many different ways. In the extreme case, $\cx$ may even contain all information in the raw state $s$, and such extraction may not help at all. Instead, we desire a fit with the \emph{minimal} $\cx$, defined as being least informative of $s$ (so that removal of the other latent variables discards the most information possible).  Concretely, we aim for a fit with least $I(\xc{\{x_t\}_{t=1}^T}; \{s_t\}_{t=1}^T \given \{a_t\}_{t=1}^T)$, the mutual information $\cx$ contains about $s$ over $T$ steps. Then from this fit, we can extract a minimal \emph{Denoised MDP} of only $\cx$.
For notation simplicity, we use bold symbols to denote variable sequences, and thus write, \eg,  $I(\cbx; \bs \given \ba)$.

Practically, we consider regularizing model-fitting with $I(\cbx; \bs \given \ba)$.
As we show below, this amounts to a modification to the well-established variational objective \citep{hafner2019dream}. The resulting method is easy-to-implement yet effective, enabling clean removal of various noise distractors the original formulation cannot handle (see \Cref{sec:expr}). 

We instantiate this idea with the structure in \Cref{fig:mdp-grid-xyz}. The \Cref{fig:mdp-grid-xy} formulation can be obtained by simply removing the  $\cz$ components and viewing $\cy$ as combined $\cyr$ and $\cyrb$.

The transition structure is modeled with components:\begin{align*}
    p_\theta^{(\cxt)}\hspace{-1pt} \trieq {}
    & p_\theta(\cxt \given \cxpt, a) \tag{$\cx$ dynamics}\\[0.2ex]
    & p_\theta(\xc{r_x} \given \cxt)  \tag{$\cx$ reward}\\[0.2ex]
    p_\theta^{(\cyt)}\hspace{-1pt} \trieq {}
    & p_\theta(\cypt \given \cypt) \tag{$\cy$ dynamics}\\[0.2ex]
    & p_\theta(\yc{r_y} \given \cyt)  \tag{$\cy$ reward}\\[0.2ex]
    p_\theta^{(\czt)}\hspace{-1pt} \trieq {}
    & p_\theta(\czt \given \cxt, \cyt, \czpt, a)  \tag{$\cz$ dynamics}\\[0.2ex]
    & p_\theta(s_t \given \cxt, \cyt, \czt) \tag{obs.~emission}\mathrlap{.}
\end{align*}
Consider training data in the form of trajectory segments $\bs, \ba, \br$ sampled from some data distribution $p_\mathsf{data}$ (\eg, stored agent experiences from a replay buffer). We perform model learning by minimizing the negative log likelihood: \begin{equation*}
    \mathcal{L}_\mathsf{MLE}(\theta) \trieq - \mathbb{E}_{\bs, \ba, \br \sim p_\mathsf{data}} \big[ \log p_\theta\left(\bs, \br \given \ba\right) \big].
\end{equation*} To obtain a tractable form, we jointly learn three variational posterior components (\ie, encoders): \begin{align*}
    q_\psi^{(\cxt)}\hspace{-1pt} \trieq {}
    & q_\psi(\cxt \given \cxpt, \cypt, \czpt, s_t, a_t) \tag{$\cx$ posterior}\\[0.2ex]
    q_\psi^{(\cyt)}\hspace{-1pt} \trieq {}
    & q_\psi(\cyt\given \cxpt, \cypt, \czpt, s_t, a_t) \tag{$\cy$ posterior}\\[0.2ex]
    q_\psi^{(\czt)}\hspace{-1pt} \trieq {}
    & q_\psi(\czt \given \cxt, \cyt, s_t, a_t) \tag{$\cz$ posterior}\mathrlap{,}
\end{align*}whose product defines the posterior $q_\psi(\cbx, \cby, \cbz \given \bs, \ba)$\footnote{Following  Dreamer \citep{hafner2019dream}, we define posterior of first-step latents $q_\psi(\xc{x_1}, \yc{y_1}, \zc{z_1} \given s_1) \trieq q_\psi(\wcdot, \wcdot, \wcdot \given \boldsymbol{0}, \boldsymbol{0}, \boldsymbol{0}, s_1, \boldsymbol{0})$, where $\boldsymbol{0}$ is the all zeros vector of appropriate size.}. We choose this factorized form based on the forward (prior) model structure of \Cref{fig:mdp-grid-xyz}.

 Then, the model can be optimized \wrt the standard variational bound on log likelihood: %
 {\small\begin{align}
    \mathcal{L}_\mathsf{MLE}(\theta) & = \min_\psi
    \underset{\bs, \ba, \br}{\mathbb{E}}
        \underset{\substack{\cbx, \cby, \cbz\sim  \\q_\psi(\cdot \given \bs, \ba, \br)}}{\mathbb{E}} \bigg[ \underbrace{- \log p_\theta(\bs, \br \given  \cbx, \cby, \cbz, \ba)}_{\textstyle \trieq \mathcal{L}_\mathsf{recon}(\theta,\psi)}  \notag\\[-0.9ex]
    & \hspace{0.1cm}{}
    + \underbrace{\sum_{t=1}^T D_\mathsf{KL}\big(
    q_\psi^{(\cxt)} \mathrel{\big\Vert} p_\theta^{(\cxt)}\big)}_{\textstyle \trieq \mathcal{L}_{\mathsf{KL}\hbox{-}\cx}(\theta,\psi)}
    + \underbrace{\sum_{t=1}^T D_\mathsf{KL}\big(
    q_\psi^{(\cyt)} \mathrel{\big\Vert} p_\theta^{(\cyt)}\big)}_{\textstyle \trieq \mathcal{L}_{\mathsf{KL}\hbox{-}\cy}(\theta,\psi)} \notag\\[-0.9ex]
    & \hspace{0.1cm}{}
    + \underbrace{\sum_{t=1}^T D_\mathsf{KL}\big(
    q_\psi^{(\czt)} \mathrel{\big\Vert} p_\theta^{(\czt)}\big)}_{\textstyle \trieq \mathcal{L}_{\mathsf{KL}\hbox{-}\cz}(\theta,\psi)} \bigg],
    \label{eq:variational-mle}
\end{align}}%
where equality is attained by optimal $q_\psi$ that is compatible with $p_\theta$, \ie, $q_\psi$ is the exact posterior of $p_\theta$.


The mutual information regularizer $I(\cbx; \bs \given \ba)$, using a variational formulation, can be written as \begin{equation}
    I(\cbx; \bs \given \ba) = \min_{\theta} \mathcal{L}_{\mathsf{KL}\hbox{-}\cx}(\theta, \psi),\label{eq:variantional-mi}
\end{equation}with equality attained when $q_\psi$ and $p_\theta$ are compatible. The appendix describes this derivation in detail.

Therefore, for a regularizer weight of $c \geq 0$, we can optimize \Cref{eq:variational-mle,eq:variantional-mi} together as \begin{align}
    &\hphantom{{}={}} \min_{\theta} \mathcal{L}_\mathsf{MLE}(\theta) + c \cdot I(\cbx; \bs \given \ba)\notag\\[-0.5ex]
    &=\min_{\theta,\psi} \mathcal{L}_\mathsf{recon}(\theta,\psi) + (1+c)\cdot\mathcal{L}_{\mathsf{KL}\hbox{-}\cx}(\theta,\psi) \notag\\[-1ex]
    &\hphantom{{}={}\min_{\theta,\psi}} + \mathcal{L}_{\mathsf{KL}\hbox{-}\cy}(\theta,\psi)  + \mathcal{L}_{\mathsf{KL}\hbox{-}\cz}(\theta,\psi). \label{eq:total-loss}
\end{align} Recall that we fit to the true MDP with the structure of \Cref{fig:mdp-grid-xyz}, which inherently guarantees all useful information in the $\cx$ latent variable. As the regularizer ensures learning the \emph{minimal} $\cx$ latents, the learned model extracts an MDP of condensed useful information with \cX as the \emph{denoised} state space, $p_\theta(\cxp \given \cx, a)$ as the transition dynamics, $p_\theta(\xc{r_x} \given \cxp)$ as the reward function. This MDP is called the \emph{Denoised MDP}, as it discards the noise distractors contained in \cy and \cz. Additionally, we also obtain $q_\psi(\cbx \given \bs, \ba)$ as the encoder mapping from raw noisy observation $s$ to the denoised $\cx$.

\myparagraph{A loss variant for improved stability.} When using a  large $c \geq 0$ (\eg when the environment is expected to be very noisy), \Cref{eq:total-loss} contains to a term with a large weight. Thus \Cref{eq:total-loss} often requires learning rates to be tuned for different $c$. To avoid this, we use the following loss form that empirically has better  training stability and does not require tuning learning rates \wrt other hyperparameters:  \begin{equation}
   \hphantom{{}={}}\min_{\theta,\psi} \mathcal{L}_\mathsf{recon} + \alpha\cdot\left(\mathcal{L}_{\mathsf{KL}\hbox{-}\cx}+ \beta\mathcal{L}_{\mathsf{KL}\hbox{-}\cy}  + \beta\mathcal{L}_{\mathsf{KL}\hbox{-}\cz}\right),\label{eq:total-loss-practice}
\end{equation}where  $\theta,\psi$ in arguments are omitted, and the hyperparameters are $\alpha > 0$ and $0 < \beta \leq 1$. Here $\beta$ is bounded, where $\beta = 1$ represents no regularization. $\alpha$ is also generally small and simply chosen according to the state-space dimensionality (see the appendix;  $\alpha \in \{1, 2\}$ in our experiments).  This form is justified from the observation that in practice we use isotropic Gaussians with fixed variance to parameterize the distributions of observation $p_\theta(s \given \dots)$ and reward $p_\theta(r \given \dots)$, where scaling log likelihoods is essentially changing the variance hyperparameter. Thus, \Cref{eq:total-loss-practice} is effectively a scaled \cref{eq:total-loss} with different variance hyperparameters. 



\begin{figure}[t]
\vspace{-0.85em}
\begin{algorithm}[H]
    \renewcommand{\algorithmicrequire}{\textbf{Input:}}
    \renewcommand{\algorithmicensure}{\textbf{Output:}}
    \small
    \caption{\small Denoised MDP}
    \label{alg:denoised-mdp}
    \begin{algorithmic}[1]
        \REQUIRE \makecell[tl]{%
        Model $p_\theta$. Posterior encoder $q_\psi$. Policy $\pi \colon \cX \rightarrow \Delta(\mathcal{A})$. \\
        Policy optimization algorithm \textsc{Pi-Opt}.
        }
        \ENSURE Denoised MDP of \cx in $p_\theta$. Encoder $q_\psi$. Policy $\pi$.
        \WHILE{training}
            \STATE {\color{BurntOrange}// Exploration}
            \STATE {Collect trajectories with $\pi$ acting on $q_\psi$ encoded outputs}
            \STATE {\color{BurntOrange}// Model learning}
            \STATE {Sample a batch of $(\bs, \ba, \br)$ segments from reply buffer}
            \STATE {Train $p_\theta$ and $q_\psi$ with \Cref{eq:total-loss-practice} on $(\bs, \ba, \br)$}
            \STATE {\color{BurntOrange}// Policy optimization}
            \STATE {Sample $\cbx \sim q_\psi(\cbx \given \bs, \ba)$; Compute $\xc{\overline{\boldsymbol{r_x}}} = \expect{p_\theta(\xc{\boldsymbol{r_x}} \given \cbx)}$}
            \STATE {Train $\pi$ by running \textsc{Pi-Opt} on $(\cbx, \ba, \xc{\overline{\boldsymbol{r_x}}})$}
        \ENDWHILE
    \end{algorithmic}
\end{algorithm}%
\vspace{-1.4em}
\end{figure}

\myparagraph{Online algorithm with policy optimization.}
The model fitting objective of \Cref{eq:total-loss-practice} can be used in various settings, \eg, offline over a collected trajectory dataset. Without assuming existing data, we explore an online setting, where the training process iteratively performs (1) exploration, (2) model-fitting, and (3) policy optimization, as shown  in \Cref{alg:denoised-mdp}. The policy $\pi \colon {\color{blue}\mathcal{X}} \rightarrow \Delta(\mathcal{A})$ soley operates on the Denoised MDP of $\cx$, which has all information sufficient for control. For policy optimization, the learned posterior encoder $q_\psi(\cbx \given \bs, \ba)$ is used to extract $\cbx$ information from the raw trajectory $(\bs, \ba, \br)$, obtaining transition sequences in $\cX$ space. Paired with the $p_\theta(\xc{\boldsymbol{r_x}} \given \cbx)$ rewards, we obtain $(\cbx, \ba, \xc{\boldsymbol{r_x}})$ as trajectories collected from the Denoised MDP on $\cx$.
Any general-purpose MDP policy optimization algorithm may be employed on these data, such as Stochastic Actor-Critic (SAC) \citep{haarnoja2018soft}.
We can also utilize the learned \emph{differentiable} Denoised MDP, \eg, optimizing policy by backpropagating through additional roll-outs from the model, as is done in Dreamer. 

While presented in the fully observable setting, Denoised MDP readily handles partial observability without extra changes. In the appendix, we discuss this point in details, and provide a guideline for choosing hyperparameters $\alpha,\beta$.


\section{Related Work}

\myparagraph{Model-Based Learning for Control} jointly learns a world model and a policy. Such methods often enjoy good sample efficiency on RL tasks with rich observations. Some formulations rely on strong assumptions, \eg, deterministic transition in DeepMDP \citep{gelada2019deepmdp} and bilinear transition in FLAMBE \citep{agarwal2020flambe}. Most general-setting methods use a reconstruction-based objective \citep{hafner2019learning,Kim2020_GameGan,ha2018world,lee2019stochastic}. Among them, Dreamer \citep{hafner2019dream} trains  world models with a variational formulation and optimizes policies by backpropagating through latent-space rollouts. It has proven effective across a variety of environments with image observations. However, such reconstruction-based approaches can struggle with the presence of noise distractors. TIA \citep{fu2021learning} partially addresses this limitation (see \Cref{sec:grid-existing}) but can not handle general distractors, unlike our method.

\myparagraph{Representation Learning and Reinforcement Learning.} Our work automates selecting useful signals from noisy MDPs by learning denoised world models, and can be viewed as an approach for general representation learning  \citep{donahue2014decaf,mikolov2013efficient,he2019momentum,huh2016makes}.
In model-free RL,
various methods learn state embeddings that are related to value functions \citep{schaul2015universal,bellemare2019geometric}, transition dynamics \citep{mahadevan2007proto,lee2020predictive}, recent action \citep{pathakICMl17curiosity}, bisimulation structure \citep{ferns2004metrics,castro2020scalable,zhang2020learning}, data augmentations \citep{laskin2020curl} \etc. 
Recently, \citet{eysenbach2021robust} proposes a regularizer similar to ours but for the different purpose of robust compressed policies. The theoretical work by \citet{efroni2021provable} is closest to our setting but concerns a more restricted set of distractors (ones both \nctrl and \nrewrel). Unlike Denoised MDP, their proposed algorithm is largely impractical and does not produce a generative model of observations (\ie, no decoder).

\myparagraph{System Identification.}
Our work is related to system identification, where an algorithm infers from real world an abstract state among a predefined limited state space, \eg,
pose estimation \citep{Guler2018DensePose,yen2020inerf} and
material  estimation \citep{hahn2019real2sim}. 
Such results are useful for
robotic manipulation \citep{manuelli2019kpam}, image generation \citep{gu2019mask}, 
\etc. Our setting is not limited to a predefined abstract state space, but instead focuses on automatic discovery of such valuable states.


\begin{figure*}[ht]
\centering
\vspace{-2.5pt}
\hspace*{-1pt}\includegraphics[scale=0.772, trim=18 65 62 25, clip]{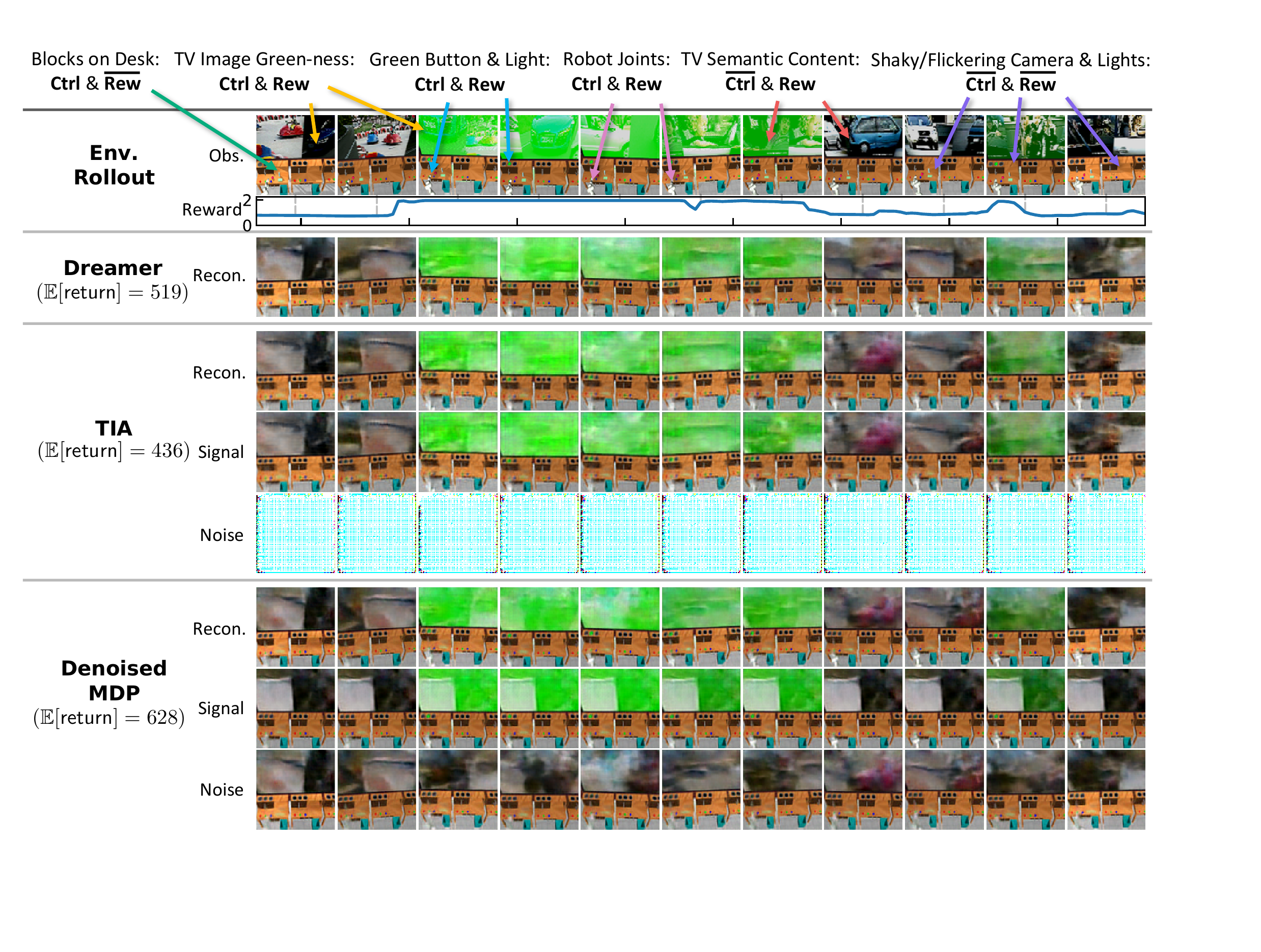}%
\vspace{-9.5pt}
\caption{Visualization of learned models for \robodesk by using decoders to reconstruct from encoded latents. For TIA and Denoised MDP, we visualize how they separate information as signal versus noise. In each row, \emph{what changes over frames is the information modeled by the corresponding latent component.} \Eg, in the bottom row, only the TV content, camera pose and lighting condition change, so Denoised MDP considers these factors as noises, while modelling the TV hue as signal. See \href{https://ssnl.github.io/denoised_mdp/\#signal-noise-factorization}{our website} for clearer video visualizations.%
}\label{fig:robo-recon}
\vspace{-7.5pt}
\end{figure*}



\section{Experiments}\label{sec:expr}

In this section, we contrast our method with existing approaches on environments with image observations and many distinct types of noise distractors. Our experiments are designed to include a variety of noise distractors and to confirm our analysis on various methods in \Cref{sec:grid-existing}.

\myparagraph{Environments.} We choose DeepMind Control (DMC) Suite \citep{tunyasuvunakool2020dm_control} (\Cref{sec:dmc}) and \robodesk \citep{kannan2021robodesk} (\Cref{sec:robodesk}) with image observations, where we explore adding various noise distractors. Information types in all evaluated environments are categorized in \Cref{table:env-info-categorize} of the appendix. Tasks include control (policy optimization) and a non-control task of regressing robot joint position from \robodesk image observations. 


\myparagraph{Methods.}
We compare not only model-based RL methods, but also model-free algorithms and general representation learning approaches, when the task is suited: \begin{itemize}
    \item \textbf{Model Learning}: Denoised MDP (our method), Dreamer \citep{hafner2019dream}, and TIA \citep{fu2021learning};
    \item \textbf{Model-Free}: DBC \citep{zhang2020learning}, \new{CURL \citep{laskin2020curl}, PI-SAC \citep{lee2020predictive} (without data augmentation for a fair comparison of its core predictive information regularization against other non-augmenting methods)}, and SAC on true state-space \citep{haarnoja2018soft} (instead of using image observations, this is roughly an ``upper bound'');
    \item \textbf{General Image Representation Learning for Non-Control Tasks}: Contrastive learning with the Alignment+Uniformity loss \citep{wang2020hypersphere} (a form of contrastive loss theoretically and empirically comparable to the popular InfoNCE loss \citep{oord2018representation}).
\end{itemize}
Model-learning methods can be used in combination with any policy optimization algorithm. For a complete comparison for general control\hide{For a more complete view of how well each model is suited for general control}, we compare the models trained with these two policy learning choices:
(1) backpropagating via the learned dynamics and (2) SAC on the learned latent space (which roughly recovers SLAC \citep{lee2019stochastic} when used with an unfactorized model such as Dreamer).

Most compared methods do not apply data augmentations, which is known to strongly boost performance \citep{yarats2021mastering,laskin2020reinforcement}. Therefore, for a fair comparison, we run PI-SAC \emph{without augmentation} to highlight its main contribution---representation of only predictive information.

All results are aggregated from 5 runs, showing mean and standard deviations. The appendix contains more details, hyperparameter studies, and additional results. Our website presents videos showing clearer video visualizations.

For Denoised MDP, we use the \Cref{fig:mdp-grid-xy} variant. Empirically, variants based on \Cref{fig:mdp-grid-xyz} lead to longer training time and sometimes inferior performance (perhaps due to having to optimize extra components and fit a more complex model). The appendix provides a comparison between them.

\subsection{\robodesk with Various Noise Distractors} \label{sec:robodesk}

We augment \robodesk environment with many noise distractors that models realistic noises (\eg, flickering lights and shaky camera). Most importantly, we place a large TV in the scene, which plays natural RGB videos. A green button on the desk controls the TV's hue (and a light on the desk). The agent is tasked with using this button to shift the TV to a green hue. Its reward is directly affected by how green the TV image is. The first row of \Cref{fig:robo-recon} shows a trajectory with various distractors annotated. All four types of information exist (see \Cref{table:env-info-categorize}), with the \ctrl and \rewrel information being the robot arm, the green button, the light on the desk, and the TV screen green-ness.

\myparagraph{Only Denoised MDP learns a clean denoised model.} Using learned decoders, \Cref{fig:robo-recon} visualizes how the models captures various information. As expected, Dreamer model captures all information.
TIA also fails to separate any noise distractors out (the \textsf{Noise} row fails to capture anything), likely due to its limited ability to model different noises. In contrast, Denoised MDP cleanly extracts all \ctrl and \rewrel information as signals---the \textsf{Signal} row only \emph{tracks changes} in robot arms, green button and light, and the TV screen green-ness. All other information is modeled as noises (see the \textsf{Noise} row). We recommend viewing video visualizations on \href{https://ssnl.github.io/denoised_mdp/\#signal-noise-factorization}{our website}.

\begin{figure*}[ht]
\centering
\includegraphics[width=1.003\linewidth, trim=3 22 15 0]{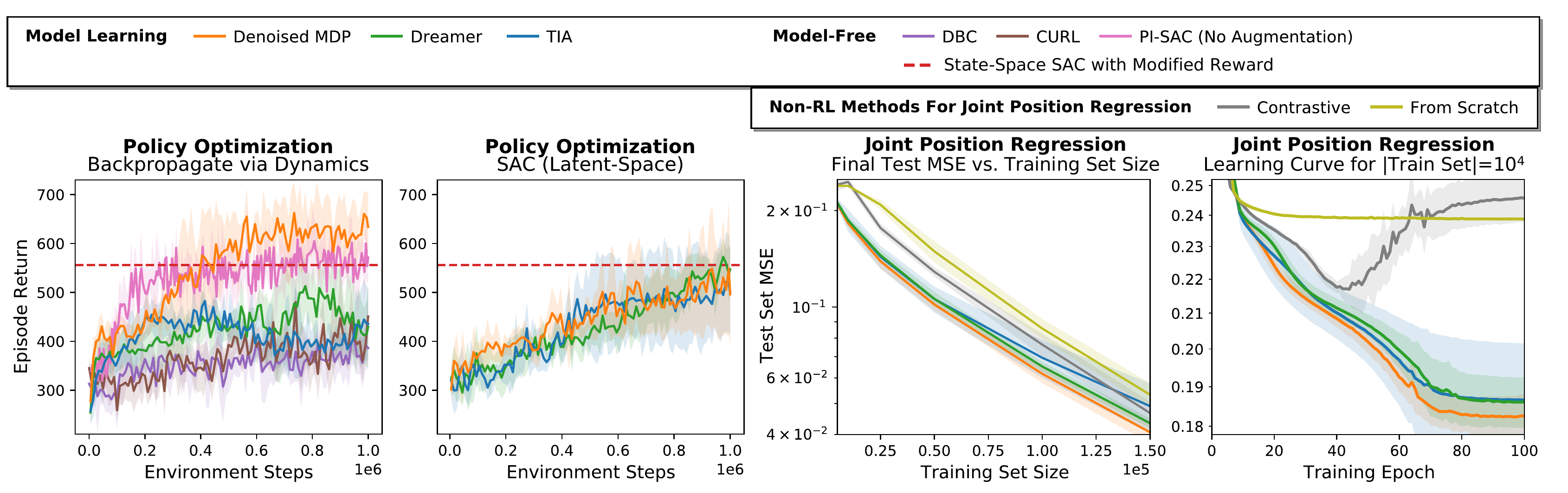}\\
\begin{minipage}[t]{0.49\linewidth}
\caption{Policy optimization on \robodesk. 
We give state-space SAC  a less noisy reward so it can learn (see appendix).
}\label{fig:robo-rew}
\end{minipage}\hfill%
\begin{minipage}[t]{0.49\linewidth}
\caption{
Performance of finetuning various encoders to infer joint position from \robodesk image observation.
}\label{fig:robo-joint}
\end{minipage}%
\vspace{-1pt}
\end{figure*}

\myparagraph{Denoised models improve policy learning.} \Cref{fig:robo-recon} also shows the total episode return achieved by policies learned with each of the three models, where the cleanest model from Denoised MDP achieves the best performance. Aggregating over $5$ runs, the complete comparison in \Cref{fig:robo-rew} shows that Denoised MDP (with backpropagating via dynamics) generally outperforms all baselines, suggesting that its clean models are helpful for control.

\begin{table*}[ht]

\vspace{-9pt}%
\hspace*{-1.2em}%
\centering%
\resizebox{
  0.815\width
}{!}{%
\small%
\centering%
\newlength{\digitl}\settowidth{\digitl}{1}
\newlength{\basel}\settowidth{\basel}{.1}
\newlength{\tinydigitl}\settowidth{\tinydigitl}{\tiny1}
\newlength{\tinybasel}\settowidth{\tinybasel}{\tiny.1}
\newcommand{\mround}[3][3]{\makebox[\basel+3\digitl][r]{\round{#2}{1}}~{\tiny\textpm~\makebox[\tinybasel+#1\tinydigitl][r]{\round{#3}{1}}}}%
\newcommand{\mroundprec}[1]{\round{#1}{2}\%\xspace}%
\newcommand{\bftab}{\fontseries{b}\selectfont}%
\renewcommand{\arraystretch}{1.3}%
\renewcommand{\arraystretch}{1.28}%
\setlength{\tabcolsep}{0.3em} 
\setlength\extrarowheight{1pt}
\begin{tabular}{ccccccccc@{\hskip 1.6em}cc}
    \toprule
    &  \multicolumn{3}{c}{\textbf{Policy Learning:} Backprop via Dynamics}
    &  \multicolumn{3}{c}{\textbf{Policy Learning:} SAC (Latent-Space)}
    & \multirow{2}{*}[-3pt]{DBC}
    & \multirow{2}{*}[-3pt]{\shortstack{PI-SAC\\[0.2ex](No Aug.)}}
    & \color{gray}\multirow{2}{*}[-3pt]{\shortstack{CURL\\[0.22ex](Use Aug.)}}
    & \color{gray}\multirow{2}{*}[-3pt]{\shortstack{State-Space SAC\\[-0.19ex](Upper Bound)}}
    \\
    \cmidrule(l{3pt}r{3pt}){2-4} \cmidrule(l{3pt}r{3pt}){5-7}

    & Denoised MDP
    & TIA
    & Dreamer
    & Denoised MDP
    & TIA
    & Dreamer
    &
    &
    &
    &
    \\
    \midrule

    \textbf{Noiseless}
    & \mround{801.4036376953126}{96.5631806020146}
    & \mround{769.6890177408854}{97.07365040080172}
    & \bftab\mround{848.5710754394531}{137.1229304081701}
    & \bftab\mround{587.0703776041667}{98.65461107426525}
    & \mround{480.2342773437501}{125.52896905820866}
    & \mround{575.4169545491537}{146.1595737300942}
    & \mround[2]{297.3831787109375}{72.5317611694336}
    & \mround[2]{246.4079319000244}{56.6048894290624}
    & \color{gray}\mround{417.27490234375}{183.1774444580078}
    & \color{gray}\mround{910.32275390625}{28.20912742614746}
    \\
    
    \textbf{Video Background}
    & \bftab\mround{597.7299418131511}{117.80636716375385}
    & \mround{407.12691319783534}{225.361746235247}
    & \mround{227.78392486572264}{102.74045289489862}
    & \mround{309.84704717000324}{153.0209530309676}
    & \bftab\mround{318.1289639790853}{123.65134941187614}
    & \mround{188.7033597310384}{78.19801845086053}
    & \mround[2]{187.9536895751953}{67.35724639892578}
    & \mround[2]{131.7182393391927}{20.061926016930414}
    & \color{gray}\mround{477.962562790126}{113.45766739421701}
    & \color{gray}\mround{910.32275390625}{28.20912742614746}
    \\
    
    \multirow{2}{*}{\shortstack{\textbf{Video Background}\\[-0.2ex]\textbf{+ Noisy Sensor}}}
    & \bftab\multirow{2}{*}{\mround{563.1350830078126}{142.9780084788406}}
    & \multirow{2}{*}{\mround{261.16744333902994}{200.36596272685793}}
    & \multirow{2}{*}{\mround{212.40484008789062}{89.70869711827993}}
    & \bftab\multirow{2}{*}{\mround{288.17919921875}{123.35275437438976}}
    & \multirow{2}{*}{\mround{197.2980753580729}{124.17133774817682}}
    & \multirow{2}{*}{\mround{218.22257232666016}{58.14234620401891}}
    & \multirow{2}{*}{\mround[2]{79.92886352539062}{35.982017517089844}}
    & \multirow{2}{*}{\mround[2]{152.49693603515627}{12.565573870135873}} 
    & \color{gray}\multirow{2}{*}{\mround{354.32098393849896}{119.85382267836293}}
    & \color{gray}\multirow{2}{*}{\mround{919.7915649414062}{100.7192153930664}}
    \\
    & & & & & & & & & &
    \\[-4.5pt]
    
    \multirow{2}{*}{\shortstack{\textbf{Video Background}\\[-0.2ex]\textbf{+ Camera Jittering}}}
    & \bftab\multirow{2}{*}{\mround{254.08687286376957}{114.17497499531767}}
    & \multirow{2}{*}{\mround{151.69740829467773}{160.50280245308895}}
    & \multirow{2}{*}{\mround{98.55416285196941}{27.744005563777023}}
    & \bftab\multirow{2}{*}{\mround{186.8494852701823}{47.67715965353988}}
    & \multirow{2}{*}{\mround{126.46733729044597}{125.60276425491514}}
    & \multirow{2}{*}{\mround{105.15849736531577}{33.79692949925169}}
    & \multirow{2}{*}{\mround[2]{68.00225067138672}{38.433292388916016}}
    & \multirow{2}{*}{\mround[2]{91.62684427897136}{7.587171023260147}}
    & \bftab\color{gray}\multirow{2}{*}{\mround{390.4128860804934}{64.91543681192832}}
    & \color{gray}\multirow{2}{*}{\mround{910.32275390625}{28.20912742614746}}
    \\
    & & & & & & & & & &
    \\[-1pt]
    \bottomrule
\end{tabular}%
}%
\vspace{-5pt}%
\caption{\new{DMC policy optimization results. For each variant, we aggregate performance across three tasks (Cheetah Run, Walker Walk, Reacher Easy) by averaging. Denoised MDP performs well across all four variants with distinct noise types. \textbf{Bold numbers} show the best model-learning result for specific policy learning choices, or the best overall result. On \textbf{Camera Jittering}, Denoised MDP greatly outperforms all other methods except for CURL, which potentially benefits from its specific data augmentation choice (random crop) on this task, and can be seen as using extra information (\ie, knowing the noise distractor form). In fact, Denoised MDP is the only method that consistently performs well across all tasks and noise variants, which can be seen from the full results in the appendix.}}\label{tbl:dmc-agg-rew}
\vspace{-3pt}

\end{table*}
\myparagraph{Denoised models benefit non-control tasks.} We evaluate the learned representations on a \emph{supervised non-control} task---regressing the robot arm joint position from observed images. Using various pretrained encoders, we finetune on a labeled training set, and measure mean squared error (MSE) on a heldout test set. In addition to RL methods, we compare encoders learned via general contrastive learning on the same amount of data. In \Cref{fig:robo-joint}, Denoised MDP representations lead to best converged solutions across a wide range of training set sizes, achieve faster training, and avoid overfitting when the training set is small. \new{DBC, CURL and PI-SAC encoders, which take in stacked frames, are not directly comparable and thus absent from \Cref{fig:robo-joint}. In the appendix, we compare them with running Denoised MDP encoder on each frame and concatenating the output features, where Denoised MDP handily outperforms both DBC and CURL by a large margin.}


\begin{figure*}[ht]
\centering
\vspace{-2pt}
\hspace*{-0.85em}
\includegraphics[trim=22 23 10 0, clip, scale=0.7925]{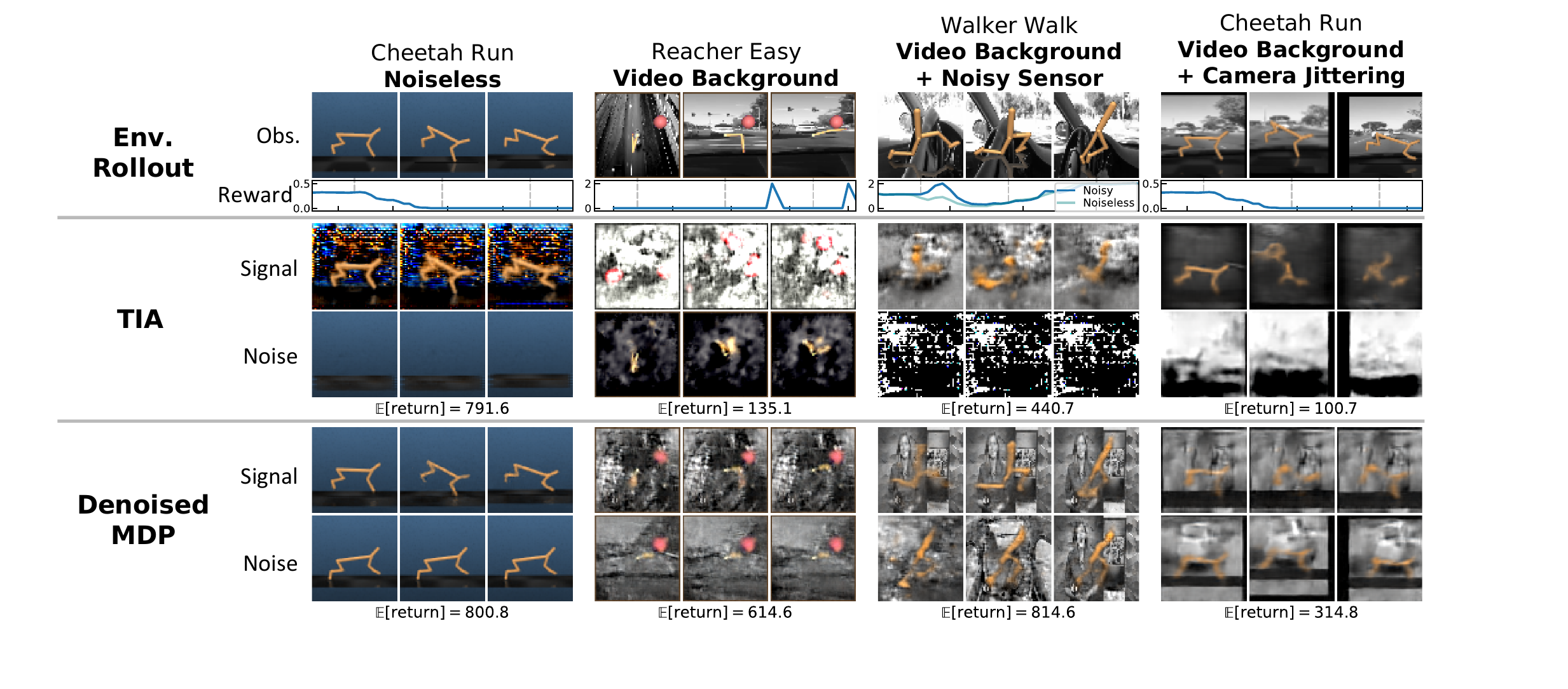}\vspace{-13pt}
\caption{Visualization of the different DMC variants  and factorizations learned by TIA and Denoised MDP. \Eg, bottom \textsf{Noise} row often shows a static agent but varying background, indicating that only the background is modeled as noises in Denoised MDP. Visualizations of full reconstructions are in appendix. See \href{https://ssnl.github.io/denoised_mdp/\#signal-noise-factorization}{our website} for clearer video visualizations.
}\label{fig:dmc-recon}
\end{figure*}

\subsection{DeepMind Control Suite (DMC)}\label{sec:dmc}

To evaluate a diverse set of noise distractors, we consider four variants for each DMC task (see \Cref{fig:dmc-recon} top row): \begin{itemize}
    \item \textbf{Noiseless}: Original environment without distractors.
    \item \textbf{Video Background}: Replacing noiseless background with natural videos \citep{zhang2020learning} ($\overbar{\textbf{Ctrl}}+\overbar{\textbf{Rew}}$).
    \item \textbf{Video Background + Sensor Noise}: Imperfect sensors sensitive to intensity of a background patch ($\overbar{\textbf{Ctrl}}+\textbf{Rew}$).
    \item \textbf{Video Background + Camera Jittering}: Shifting the observation by a smooth random walk ($\overbar{\textbf{Ctrl}}+\overbar{\textbf{Rew}}$).
\end{itemize}

\myparagraph{Denoised MDP consistently removes noise distractors.}In \Cref{fig:dmc-recon}, TIA struggles to learn clean separations in many settings. Consistent with analysis in \Cref{sec:grid-existing}, it cannot handle \textbf{Sensor Noise} or \textbf{Camera Jittering}, as the former is \rewrel noise that it cannot model, and the latter (although \nrewrel) cannot be represented by masking. Furthermore, it fails on Reacher Easy with \textbf{Video Background}, where the reward is given by the distance between the agent and a randomly-located ball. TIA encourages its noise latent to be independent of reward, but does not prevent it from capturing the \ctrl agent. These failures lead to either TIA trying to model everything as useful signals, or a badly-fit model (\eg, wrong agent pose in the last column). In contrast, Denoised MDP separates out noise in all cases, obtaining a clean and accurate MDP (its \textsf{Signal} rows only have the agent moving).


\myparagraph{Denoised models consistently improve policy learning. } We evaluate the learned policies in \Cref{tbl:dmc-agg-rew}, where results are aggregated by the noise distractor variant. \new{Other methods, while sometimes handling certain noise types well, struggle to deal with all four distinct variants. TIA, as expected, greatly underperforms Denoised MDP under \textbf{Noisy Sensor} or \textbf{Camera Jittering}. CURL, whose augmentation choice potentially helps handling \textbf{Camera Jittering}, underperforms in other three variants. In contrast, Denoised MDP policies \emph{consistently} perform well for all noisy variants and also the noiseless setting, regardless of the policy optimizer.}

Model-based approaches have a significant lead over the model-free ones, as seen from the DBC results in \Cref{tbl:dmc-agg-rew} and the well-known fact that direct model-free learning on raw image observations usually fails \citep{laskin2020reinforcement,kostrikov2020image,yarats2021mastering}. These results show that learning in a world model is useful, and that learning in a denoised world model is even better.




\section{Implications}

In this work we explore learning denoised and compressed world models in the presence of environment noises.

As a step towards better understanding of such noises, we categorize of information in the wild into four types (\Cref{sec:information-world}). This provides a framework to contrast and understand various methods, highlighting where they may be successful and where they will suffer (\Cref{sec:grid-existing}). Insights gained this way empirically agrees with findings from extensive experiments (\Cref{sec:expr}). It can potentially assist better algorithm design and analysis of new MDP representation methods, as we have done in designing Denoised MDP (\Cref{sec:denoise}). We believe that this categorization will be a useful framework for investigation on learning under noises, revealing not just the (conceptual) success scenarios, but also the failure scenarios at the same time. Additionally, the framework can be readily extended with more sophisticated factorizations (\Cref{sec:factorization-extension}), which can lead to corresponding Denoised MDP variants and/or new algorithms.

Based on the framework, our proposed Denoised MDP novelly can remove \emph{all} noise distractors that are \emph{\nctrl or \nrewrel}, in distinction to prior works. Empirically, it effectively identifies and removes a diverse set of noise types, obtaining clean denoised world models (\Cref{sec:expr}). It may serve as an important step towards efficient learning of general tasks in the noisy real world. Our experiments also highlight benefits of cleanly denoised world models on both standard control tasks as well as non-control tasks. The success in both cases highlights the general usefulness of such models. Given the generality of MDPs, this opens up the possibility of casting non-RL tasks as MDPs and automatically learn representations from denoised world models, as an alternative to manual feature engineering.






\section*{Acknowledgements}
We thank Jiaxi Chen for the beautiful \Cref{fig:intro-sunlight-2x2-example} illustration.
We thank Daniel Jiang and Yen-Chen Lin for their helpful comments and suggestions. We are grateful to the following organizations for providing computation resources to this project: IBM's MIT Satori cluster, MIT Supercloud cluster, and Google Cloud Computing with credits gifted by Google to MIT. We are very thankful to Alex Lamb for suggestions and catching our typo in the conditioning of \Cref{eq:variational-mle}.
\clearpage\newpage

\bibliography{reference}
\bibliographystyle{icml2022}

\newpage

\appendix
\onecolumn

\section{Denoised MDP Discussions}

\subsection{Loss Derivation}

To apply our  mutual information regularizer $I(\cbx; \bs \given \ba)$, we can consider a form using another variational distribution $\rho$ (see, \eg, \citet{poole2019variational}), \begin{align}
    I(\cbx; \bs \given \ba) \notag
    & = \min_{\rho} \mathbb{E}_{\ba} \mathbb{E}_{p_\theta(\bs \given \ba)} \left[ D_\mathsf{KL}(p_\theta(\cbx \given \bs, \ba) \mathrel{\Vert} \rho(\cbx \given \ba)) \right] \notag\\
    & \approx \min_{\rho} \mathbb{E}_{\ba} \mathbb{E}_{q_\psi(\bs \given \ba)} \left[ D_\mathsf{KL}(q_\psi(\cbx \given \bs, \ba) \mathrel{\Vert} \rho(\cbx \given \ba)) \right] \tag{assume $q_\psi$ is roughly the posterior of $p_\theta$}\\
    & = \min_{\theta'} \mathcal{L}_{\mathsf{KL}\hbox{-}\cx}(\psi, \theta'). \label{eq:appendix-variational-mi}
\end{align}
The assumption that $q_\psi$ is roughly the posterior of $p_\theta$ is acceptable because it is the natural consequence of optimizing the variational MLE objective in \Cref{eq:variational-mle} over $\theta, \psi$.

Alternatively, we can consider the MI defined by a joint conditional distribution $P(\cbx, \bs \given a)$ not from the forward model $p_\theta$, but from the data distribution and posterior model $q_\psi(\cbx \given \bs, \ba)$. This is also sensible because the variational MLE objective in \Cref{eq:variational-mle} optimizes for compatible $p_\theta$ and $q_\psi$ that both fit data and consistently describe (conditionals of) the same underlying distribution. Thus regularizing either can encourage a low MI. This approach leads to exactly \Cref{eq:appendix-variational-mi}, without approximation.

Then, the total loss in \Cref{eq:total-loss} from combining \Cref{eq:variational-mle,eq:appendix-variational-mi} is given by \begin{align}
    \min_{\theta} \mathcal{L}_\mathsf{MLE}(\theta) + c \cdot I(\cbx; \bs \given \ba)
    & =
    \min_{\theta, \theta', \psi}  \mathcal{L}_\mathsf{recon}(\theta,\psi)
    + \mathcal{L}_{\mathsf{KL}\hbox{-}\cx}(\theta,\psi)  + \mathcal{L}_{\mathsf{KL}\hbox{-}\cy}(\theta,\psi)  + \mathcal{L}_{\mathsf{KL}\hbox{-}\cz}
    + c \cdot  + \mathcal{L}_{\mathsf{KL}\hbox{-}\cx}(\theta',\psi)  \notag \\
    &=\min_{\theta,\psi} \mathcal{L}_\mathsf{recon}(\theta,\psi) + (1+c)\cdot\mathcal{L}_{\mathsf{KL}\hbox{-}\cx}(\theta,\psi)  + \mathcal{L}_{\mathsf{KL}\hbox{-}\cy}(\theta,\psi)  + \mathcal{L}_{\mathsf{KL}\hbox{-}\cz}(\theta,\psi). \notag
\end{align}

\subsection{Discussions}\label{sec:appendix-denoised-mdp-discussions}

We discuss some algorithmic choices of Denoised MDP below. Specific implementation details (\eg, architectures) can be found at \Cref{sec:appendix-model-learning-details}.

\myparagraph{Posterior distributions of $\xc{r_x}$ and $\yc{r_y}$.}
The $p_\theta$ reward distributions $p_\theta(\xc{r_x} \given \cxt)$ and $p_\theta(\yc{r_y} \given \cyt)$ are modelled via Gaussians (as is done usually in world models, such as Dreamer \citep{hafner2019dream}). By the transition structure of Denoised MDPs, these distributions are inherently independent. Recall that $r = \xc{r_x} + \yc{r_y}$. Therefore, we can easily compute the distribution of $p_\theta({r} \given \cxt, \cyt)$ and its log likelihoods. This enables easy optimization of the variational MLE objective, without requiring the posterior model to also infer $\xc{r_x}$ and $\yc{r_y}$ from observed $r$ subject to the addition relation.

\myparagraph{Partial observability.}\Cref{sec:information-world,sec:denoise} discussions are mostly based in the fully observable setting. Yet most benchmarks and real-world tasks are partially observable, \eg, robot joint speeds that can not be inferred from a single frame. Fortunately, the transition models used in Denoised MDP are fully capable of handle such cases, as long as the encoder $q_\psi$ is not deterministic and the observation model $p_\theta(s \given \dots)$ does not have the block structure \citep{du2019provably} (which would make $\cx, \cy, \cz$ fully determined from $s$). In practice, we let both components to be generic conditional distributions (parameterized by regular deep neural networks). Therefore, Denoised MDP does not require full observability.

\myparagraph{Hyperparameter choice.} The loss in \Cref{eq:total-loss-practice} has two hyperparameters: $\alpha \in (0, \infty)$ and $\beta \in (0, 1)$. To maintain relative ratio with the observation reconstruction loss, we recommend scaling $\alpha$ roughly proportionally with dimensionality of the observation space, as is done in our experiments presented in this paper. A smaller $\beta$ means stronger regularization. Therefore, $\beta$ can be chosen based on training stability and the level of noise distractors in the task.


\section{Experiment Details}

All code (including code for our environment variants and code for our Denoised MDP method) will be released upon publication.

\subsection{Implementation Details}

\subsubsection{Environments and Tasks}

\begingroup
\newcommand{\NA}{---}
\renewcommand{\arraystretch}{1.75}
\begin{table}
\centering
\resizebox{
  0.89\width
}{!}{%
\small
\begin{tabular}{cccccc}
    \toprule
    \multicolumn{2}{c}{}
    & $\textbf{Ctrl}+\textbf{Rew}$
    & $\textbf{Ctrl}+\overbar{\textbf{Rew}}$
    & $\overbar{\textbf{Ctrl}}+\textbf{Rew}$
    & $\overbar{\textbf{Ctrl}}+\overbar{\textbf{Rew}}$
    \\
    \midrule

    \multirow{6}{*}{DMC}
    & \textbf{Noiseless}
    & Agent
    & \NA
    & \NA
    & \NA
    \\

    & \textbf{Video Background}
    & Agent
    & \NA
    & \NA
    & Background
    \\

    & \multirow{2}{*}{\shortstack{\textbf{Video Background}\\[-0.2ex]\textbf{+ Noisy Sensor}}}
    & \multirow{2}{*}{Agent}
    & \multirow{2}{*}{\NA}
    & \multirow{2}{*}{Background}
    & \multirow{2}{*}{\NA}
    \\
    &
    &
    &
    &
    &
    \\

    & \multirow{2}{*}{\shortstack{\textbf{Video Background}\\[-0.2ex]\textbf{+ Camera Jittering}}}
    & \multirow{2}{*}{Agent}
    & \multirow{2}{*}{\NA}
    & \multirow{2}{*}{\NA}
    & \multirow{2}{*}{\shortstack{Background,\\Jittering camera}}
    \\
    &
    &
    &
    &
    &
    \\
    \midrule

    \multicolumn{2}{c}{\multirow{3}{*}{\robodesk}}
    & \multirow{3}{*}{\shortstack{Agent, Button,\\Light on desk,\\Green hue of TV}}
    & \multirow{3}{*}{\shortstack{Blocks on desk,\\Handle on desk,\\Other movable objects }}
    & \multirow{3}{*}{\shortstack{TV content,\\Button sensor noise}}
    & \multirow{3}{*}{\shortstack{Jittering and flickering environment lighting,\\Jittering camera}}
    \\

    \multicolumn{2}{c}{}
    &
    &
    &
    &
    \\

    \multicolumn{2}{c}{}
    &
    &
    &
    &
    \\
    \bottomrule
\end{tabular}%
}
\caption{Categorization of various information in the environments we evaluated with.}\label{table:env-info-categorize}
\end{table}
\endgroup

In all environments, trajectories are capped at $1000$ timesteps. \Cref{table:env-info-categorize} shows a summary of what kinds of information exist in each environment.

\paragraph{DeepMind Control Suite (DMC).} Our \textbf{Video Background} implementation follows Deep Bisimulation for Control \citep{zhang2020learning} on most environments, using \verb|Kinetics-400| grayscale videos \citep{smaira2020short}, and  replacing pixels where blue channel is strictly the greatest of three. This method, however, does not cleanly remove most of background in the Walker Walk environment, where we use an improved mask that replaces all pixels where the blue channel is \emph{among the greatest} of three. For \textbf{Camera Jittering}, we shift the observation image according to a smooth random walk, implemented as, at each step, Gaussian-perturbing acceleration, decaying velocity, and adding a pulling force if the position is too far away from origin. For \textbf{Sensor Noise}, we select one sensor, and perturb it according to intensity of a patch of the natural video background (\ie, adding average patch value ${} - 0.5$). We perturb the \verb|speed| sensor for Cheetah Run, the \verb|torso_height| sensor for Walker Walk, and the normalized \verb|finger_to_target_dist| sensor for Reacher Easy. These sensor values undergo non-linear (mostly piece-wise linear) transforms to compute rewards. While they can not be perfectly modelled by additive reward noise, such a model is usually sufficient in most cases when the sensor values are not too extreme and stay in one linear region.

\paragraph{\robodesk.} We modify the original \robodesk environment by adding a TV screen and two neighboring desks. The TV screen places (continuously horizontally shifting) natural RGB videos from the  \verb|Kinetics-400| dataset \citep{smaira2020short}. The environment has three light sources from the above, to which we added random jittering and flickering. The viewing camera is placed further to allow better view of the noise distractors. Resolution is increased from $64\times 64$ to $96 \times 96$ to compensate this change. Camera jittering is implemented by a 3D smooth random walk. Finally, the button sensor (\ie, detected value of how much the button is pressed) is also offset by a random walk. Each of the three button affects the corresponding light on the desk. Additionally, pressing the green button also shifts the TV screen content to a green hue. Following \robodesk reward design, we reward the agent for (1) placing arm close to the button, (2) pressing the button, and (3) how green the TV screen content is.

\paragraph{\robodesk Joint Position Regression Datasets.}To generate training and test set, we use four policies trained by state-space SAC at different stages of training (which is not related to any of the compared methods) and a uniform random actor, to obtain five policies of different qualities. For each policy, we sample $100$ trajectories, each containing $1001$ pairs (from $1000$ interactions) of image observation and groundtruth joint position (of dimension $9$). This leads to a total of $500.5 \times 10^3$ samples from each policy. From these, $100 \times 10^3$ samples are randomly selected as test set. Training sets of sizes $5 \times 10^3, 10 \times 10^3, 25 \times 10^3, 50 \times 10^3, 100 \times 10^3, 150 \times 10^3$ are sampled from the rest. For all test sets and training sets, we enforce each policy to strictly contribute an equal amount.

\subsubsection{Model Learning Methods}\label{sec:appendix-model-learning-details}

For all experiments, we let the algorithms use $10^6$ environment steps. For PI-SAC and CURL, we follow the original implementations \citep{laskin2020curl,lee2020predictive} and use an action repeat of $4$ for Cheetah Run and Reacher Easy, and an action repeat of $2$ for Walker Walk. For Denoised MDP, Dreamer, TIA and DBC, we always use an action repeat of $2$, following prior works \citep{hafner2019dream,fu2021learning,zhang2020learning}.


\begingroup
\newcommand{\NA}{---}
\renewcommand{\arraystretch}{1.4}
\begin{table}
\begin{minipage}[t]{0.45\linewidth}
\centering
\resizebox{
  0.89\width
}{!}{%
\small
\begin{tabular}{ccccc}
    \toprule
    \multirow{2}{*}{Operator} &
    \multirow{2}{*}{\shortstack{Input\\Shape}} &
    \multirow{2}{*}{\shortstack{Kernel\\Size}} &
    \multirow{2}{*}{Stride} &
    \multirow{2}{*}{Padding} \\

    & & & &  \\
    \midrule

    Input &
    $[3, 96, 96]$ &
    \NA &
    \NA &
    \NA  \\

    Conv. + ReLU &
    $[k, 47, 47]$&
    4 &
    2 &
    0 \\

    Conv. + ReLU &
    $[2k, 22, 22]$&
    4 &
    2 &
    0 \\

    Conv. + ReLU &
    $[4k, 10, 10]$&
    4 &
    2 &
    0 \\

    Conv. + ReLU &
    $[8k, 4, 4]$&
    4 &
    2 &
    0 \\

    Conv. + ReLU &
    $[8k, 2, 2]$&
    3 &
    1 &
    0 \\

    Reshape + FC &
    $[m]$&
    \NA &
    \NA &
    \NA \\
    \bottomrule
\end{tabular}%
}
\caption{Encoder architecture for $(96 \times 96)$-resolution observation. The output of this encoder is then fed to other network for inferring posteriors. $m$ and $k$ are two architectural hyperparameters. $m$ controls the output size (unrelated to the actual latent variable sizes). $k$ controls the network width.  }\label{table:model-based-96-enc}%
\end{minipage}\hfill%
\begin{minipage}[t]{0.53\linewidth}
\centering
\resizebox{
  0.89\width
}{!}{%
\small
\begin{tabular}{ccccc}
    \toprule
    \multirow{2}{*}{Operator} &
    \multirow{2}{*}{\shortstack{Input\\Shape}} &
    \multirow{2}{*}{\shortstack{Kernel\\Size}} &
    \multirow{2}{*}{Stride} &
    \multirow{2}{*}{Padding} \\

    & & & &  \\
    \midrule

    Input &
    $[\texttt{input\_size}]$ &
    \NA &
    \NA &
    \NA  \\

    FC + ReLU + Reshape &
    $[m, 1, 1]$ &
    \NA &
    \NA &
    \NA  \\

    Conv.~Transpose + ReLU &
    $[4k, 3, 3]$&
    5 &
    2 &
    0 \\

    Conv.~Transpose + ReLU &
    $[4k, 9, 9]$ &
    5 &
    2 &
    0 \\

    Conv.~Transpose + ReLU &
    $[2k, 21, 21]$ &
    5 &
    2 &
    0 \\

    Conv.~Transpose + ReLU &
    $[k, 46, 46]$ &
    6 &
    2 &
    0 \\

    Conv.~Transpose + ReLU &
    $[3, 96, 96]$ &
    6 &
    2 &
    0 \\
    \bottomrule
\end{tabular}%
}
\caption{Decoder architecture for $(96 \times 96)$-resolution observation. $m$ and $k$ are two architectural hyperparameters. $m$ controls width the fully connected part. $k$ controls width of the convolutional part. They are the same values as in \Cref{table:model-based-96-enc}.}\label{table:model-based-96-dec}%
\end{minipage}
\end{table}
\endgroup

\paragraph{Denoised MDP, Dreamer, and TIA.}
Both Dreamer and TIA use the same training schedule and the Recurrent State-Space Model (RSSM) as the base architecture \citep{hafner2019learning}. Following them, Denoised MDP also uses these components, and follow the same prefilling and training schedule (see Dreamer \citep{hafner2019learning} for details). These three model learning methods take in $64\times 64$ RGB observations for DMC, and $96\times 96$ RGB observations for \robodesk. Dreamer only implements encoder and decoder for the former resolution. To handle the increased resolution, we modify the $64\times 64$ architectures and obtain convolutional encoder and decoder shown in \Cref{table:model-based-96-enc,table:model-based-96-dec}. For fair comparison, we ensure that each method has roughly equal number of parameters by using different latent variable sizes, encoder output sizes ($m$ of \Cref{table:model-based-96-enc}) and convolutional net widths ($k$ of \Cref{table:model-based-96-dec}). Details are shown in \Cref{table:model-based-sizes}.

\paragraph{KL clipping (free nats).}
For Denoised MDP, we follow Dreamer \citep{hafner2019learning,hafner2019dream} and TIA \citep{fu2021learning}, and allow $3$ free nats for the $\mathcal{L}_{\mathsf{KL}\hbox{-}\cx}$ term. In other words, for each element of a batch, we do not optimize the KL term if it is less than $3$ (\eg, implemented via clipping).  However, we do not allow this for the $\mathcal{L}_{\mathsf{KL}\hbox{-}\cy}$ and $\mathcal{L}_{\mathsf{KL}\hbox{-}\cz}$ terms, as these variables are to be discarded and information is not allowed to hide in them unless permitted by the structure.  An alternative strategy, which we find also empirically effective, is to consider $\mathcal{L}_{\mathsf{KL}\hbox{-}\cx} = \underbrace{\beta\cdot \mathcal{L}_{\mathsf{KL}\hbox{-}\cx}}_{\text{VAE KL term}} + \underbrace{(1-\beta)\cdot \mathcal{L}_{\mathsf{KL}\hbox{-}\cx}}_{\text{MI regularizer term}}$, and to allow free nats only for the first term that is a part of the variational model fitting objective. All results presented in this paper use the first strategy. Both strategies are implemented in our open source code repository: \href{https://github.com/facebookresearch/denoised_mdp/}{\texttt{github.com/facebookresearch/denoised\_mdp}}.


\begingroup
\newcommand{\NA}{---}
\renewcommand{\arraystretch}{1.2}
\begin{table}
\centering
\resizebox{
  0.92\width
}{!}{%
\small
\begin{tabular}{ccccccccc}
    \toprule
    \multirow{3}{*}[-3pt]{} &
    \multicolumn{4}{c}{DMC} &
    \multicolumn{4}{c}{\robodesk} \\
    \cmidrule(lr){2-5} \cmidrule(lr){6-9}

    & \multirow{2}{*}[-2pt]{Latent Sizes}
    & \multirow{2}{*}[-2pt]{$m$}
    & \multirow{2}{*}[-2pt]{$k$}
    & \multirow{2}{*}[-2pt]{\shortstack{Total Number\\of Parameters}}
    & \multirow{2}{*}{Latent Sizes}
    & \multirow{2}{*}[-2pt]{$m$}
    & \multirow{2}{*}[-2pt]{$k$}
    & \multirow{2}{*}[-2pt]{\shortstack{Total Number\\of Parameters}}
    \\

    &
    &
    &
    &
    &
    &
    &
    &
    \\
    \midrule

    Dreamer
    & $(220+33)$
    & $1024$
    & $32$
    & $7{,}479{,}789$
    & $(220+33)$
    & $1024$
    & $32$
    & $6{,}385{,}511$
    \\

    TIA
    & $(120+20)+(120+20)$
    & $490$
    & $24$
    & $7{,}475{,}567$
    & $(120+20)+(120+20)$
    & $490$
    & $24$
    & $6{,}384{,}477$ \\

    Denoised MDP
    & $(120+20)+(120+20)$
    & $1024$
    & $32$
    & $7{,}478{,}826$
    & $(120+20)+(120+20)$
    & $1024$
    & $32$
    & $6{,}384{,}248$
     \\
    \bottomrule
\end{tabular}%
}
\caption{The specific architecture parameters for model learning methods. Since RSSM uses a deterministic part and a stochastic part to represent each latent variable, we use $(\texttt{deterministic\_size} + \texttt{stochastic\_size})$ to indicate size of a latent variable. TIA and Denoised MDP have more than one latent variable. Note that while TIA has lower $m$ and $k$, it has multiple encoder and decoders, whereas Dreamer and Denoised MDP only have one encoder and one decoder. The total number of parameters is measured with the actor model, but without any additional components from policy optimization algorithm (\eg, critics in SAC). Total number of parameters is lower for \robodesk as the encoder and decoder architecture is narrower than those of DMC for the purpose of reducing memory usage, despite with a higher resolution. }\label{table:model-based-sizes}%
\end{table}
\endgroup

\subsubsection{Policy Optimization Algorithms Used with Model Learning}

\paragraph{Backpropagate via Dynamics.} We use the same setting as Dreamer \citep{hafner2019dream}, optimizing a $\lambda$-return over $15$-step-long rollouts with $\lambda=0.95$, clipping gradients with norm greater than $100$.  TIA uses the same strategy, except that it groups different models together for gradient clipping. We strictly follow the official TIA implementation.

\paragraph{Latent-Space SAC.} We use the regular  SAC  with automatic entropy tuning, without gradient clipping. This works well for almost all settings, except for Walker Walk variant of DMC, where training often collapses after obtaining good return, regardless of the model learning algorithm. To address instability in this case, we reduce learning rates from $3\times 10^{-4}$ to $1\times 10^{-4}$ and clip gradients with norm greater than $100$ for all latent-space SAC run on these variants.

\subsubsection{Model-Free Methods}

\paragraph{DBC.} For DMC, we used $84 \times 84$-resolution observation following original work (even though other methods train on $64 \times 64$-resolution observations). For \robodesk, DBC uses the encoder in \Cref{table:model-based-96-enc} for $96\times 96$-resolution observation, for fair comparison with other methods. Following the original work, we stack $3$ consecutive frames to approximate the required full observability. In the robot arm joint position regression experiment \Cref{sec:robodesk}, DBC encoders also see stacked observations. For DMC evaluations, we use the data provided by \citeauthor{zhang2020learning} wherever possible, and run the official repository for other cases.

\paragraph{State-Space SAC.} The state space usually contains robot joint states, including position, velocity, \etc.  For DMC, when \textbf{Sensor Noise} is present, this is not the true optimal state space, as we do not supply it with the noisy background that affects the noisy reward. However, it still works well in practice. For \robodesk, the TV's effect on reward is likely stronger and direct state-space SAC fails to learn. Since this evaluation is to obtain a rough ``upper bound'', we train state-space SAC with a modified  reward with less noise--- the agent is rewarded by pressing the button, independent of the TV content. This  still encourages the optimal strategy of the task allows achieving good policies.

\subsubsection{Non-RL methods}

\paragraph{Contrastive Learning.} We used the Alignment+Uniformity contrastive learning loss from \citet{wang2020hypersphere}. The hyperparameters and data augmentations strictly follow their experiments on STL{-}10 \citep{coates2011stl10}, which also is of resolution $96\times 96$. The exact loss form is $\mathcal{L}_\mathsf{align}(\alpha=2) + \mathcal{L}_\mathsf{uniform}(t=2)$, a high-performance setting for STL{-}10.

\subsection{Compute Resources}\label{sec:compute-resource}

All our experiments are run on a single GPU, requiring 8GB memory for DMC tasks, and 16GB memory for \robodesk tasks.  We use NVIDIA GPUs of the following types: 1080 Ti, 2080 Ti, 3080 Ti, P100, V100, Titan XP, Titan RTX. For \verb|MuJoCo| \citep{todorov2012mujoco}, we use the \verb|EGL| rendering engine. Training time required for each run heavily depends on the CPU specification and availability. In general, a Denoised MDP run needs $12\sim36$ hours on DMC and $24\sim 50$ hours on \robodesk. TIA uses about $1.5\times$ of these times, due to the adversarial losses. For a comparison between the two Denoised MDP variants, running the same DMC task on the same machine, the \Cref{fig:mdp-grid-xy} variant used $23$ hours while variants based on \Cref{fig:mdp-grid-xyz} used $26$ hours. See \Cref{sec:appendix-dmc-details} for details on \Cref{fig:mdp-grid-xyz} details.


\subsection{Visualization Details}

\paragraph{Visualizations of components in learned models.}
We use different methods to visualize signal and noise information learned by TIA and Denoised MDP in \Cref{fig:dmc-recon,fig:robo-recon}. For TIA, we used the reconstructions from the two latent (before mask-composing them together as the full reconstruction). For Denoised MDP, we only have one decoder (instead of three for TIA), and thus we decode $(\xc{x_t}, \texttt{const})$ and $(\texttt{const}, \yc{y_t})$ to visualize information contained in each variable, with $\texttt{const}$ chosen by visual clarity (usually as value of the other variable at a fixed timestep). Due to the fundamental different ways to obtain these visualizations, in DMC, TIA can prevent the agent from showing up in noise visualizations, while Denoised MDP cannot. However, as stated in \Cref{sec:dmc}, our focus should be on what evolves/changes in these images, rather than what is visually present, as static components are essentially not modelled by the corresponding transition dynamics.
Visualizations in \Cref{fig:robo-recon,fig:dmc-recon} use trajectories generated by a policy trained with state-space SAC. To obtain diverse behaviors, policy outputs are randomly perturbed before being used as actions. From the same trajectory, we use the above described procedure to obtain visualizations.  The specific used trajectory segments are chosen to showcase both the modified environment and representative behavior of each method. Please see the supplementary video for clearer visualizations.

\subsection{\robodesk Result Details}

\paragraph{Environment modifications.} The agent controls a robotic arm placed in front of a desk and a TV, and is tasked to push down the green button on the desk, which turns on a small green light and makes the TV display have a green hue. The intensity of the TV image's green channel is given to the agent as part of their reward, in addition to distance between the arm to the button, and how much the button is pressed. Additionally, the environment contains other noise distractors, including moveable blocks on the desk ($\textbf{Ctrl}+\overbar{\textbf{Rew}}$), flickering environment light and camera jittering  ($\overbar{\textbf{Ctrl}}+\overbar{\textbf{Rew}}$), TV screen hue   ($\textbf{Ctrl}+\textbf{Rew}$), TV content  ($\overbar{\textbf{Ctrl}}+\textbf{Rew}$), and noisy button sensors  ($\overbar{\textbf{Ctrl}}+\textbf{Rew}$).

\paragraph{Denoised MDP hyperparameters.} \robodesk has roughly twice as many pixels as DMC has. For Denoised MDP, we  scale $\alpha$ with the observation space dimensionality (see \Cref{sec:denoise}) and use $\alpha=2$, with a fixed $\beta=0.125$. When using the alternative KL free nats strategy discussed in \Cref{sec:appendix-model-learning-details} (results not shown in paper), we find $\alpha=1$ and $\beta=0.25$ also effective.

\paragraph{TIA hyperparameters.} We follow recommendations in the TIA paper, setting $\lambda_\text{Radv}=25{,}000$ to match reconstruction loss in magnitude, and setting $\lambda_{O_s}=2$ where training is stable.


\subsubsection{Robot Arm Joint Position Regression.}

\begin{figure}[t]
    \centering
    \begin{minipage}[t]{0.5\textwidth}
    \centering
    \includegraphics[width=\linewidth, trim=20 20 0 0]{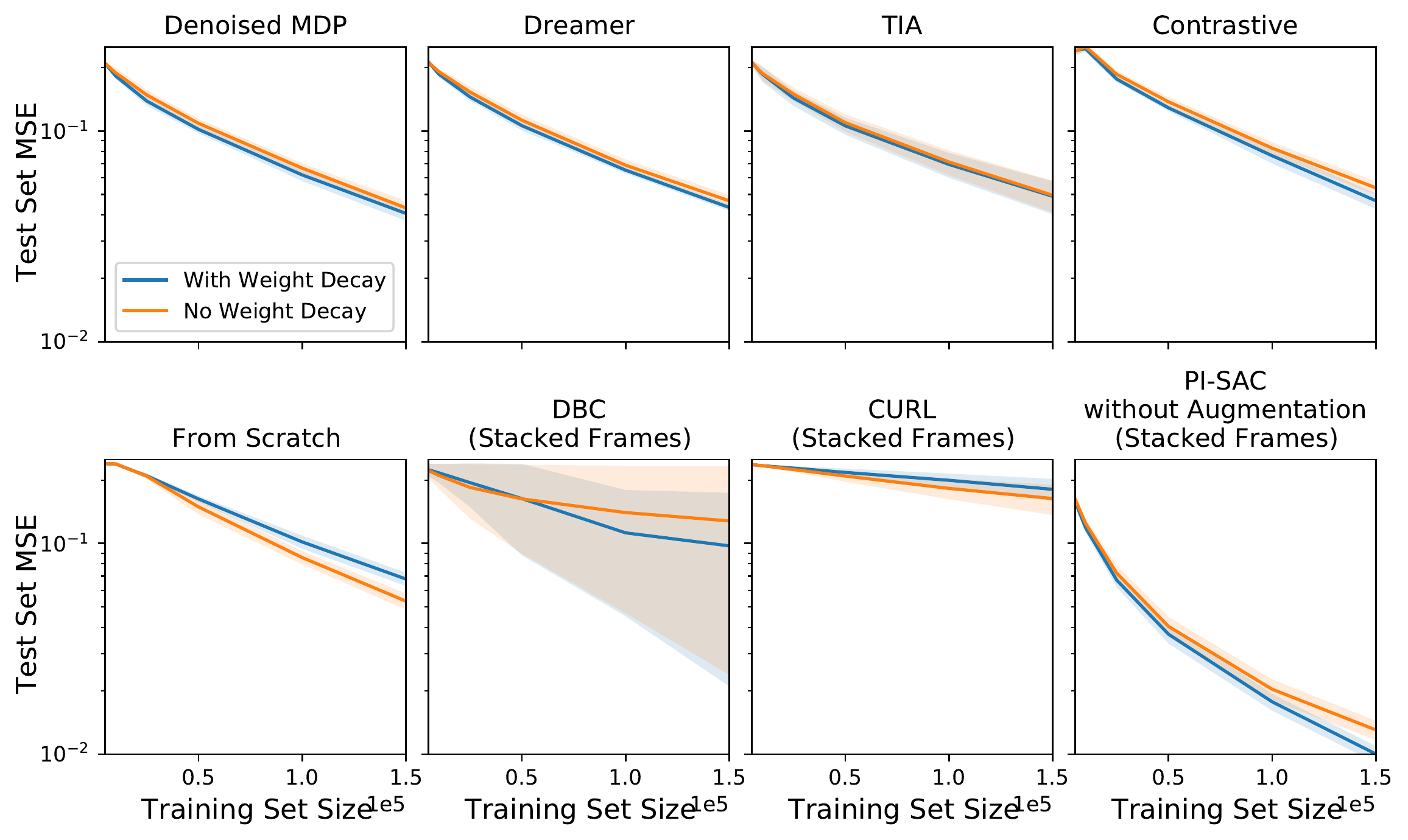}
    \caption{Effect of weight decay on \robodesk joint position regression. The curves show final test MSE for various training set sizes. Weight decay generally helps when finetuning from a pretrained encoder, but hurts when training from scratch.}
    \label{fig:robo-joint-wd}
    \end{minipage}\hfill%
    \begin{minipage}[t]{0.48\textwidth}
    \centering
    \includegraphics[width=0.9\linewidth, trim=0 20 0 0]{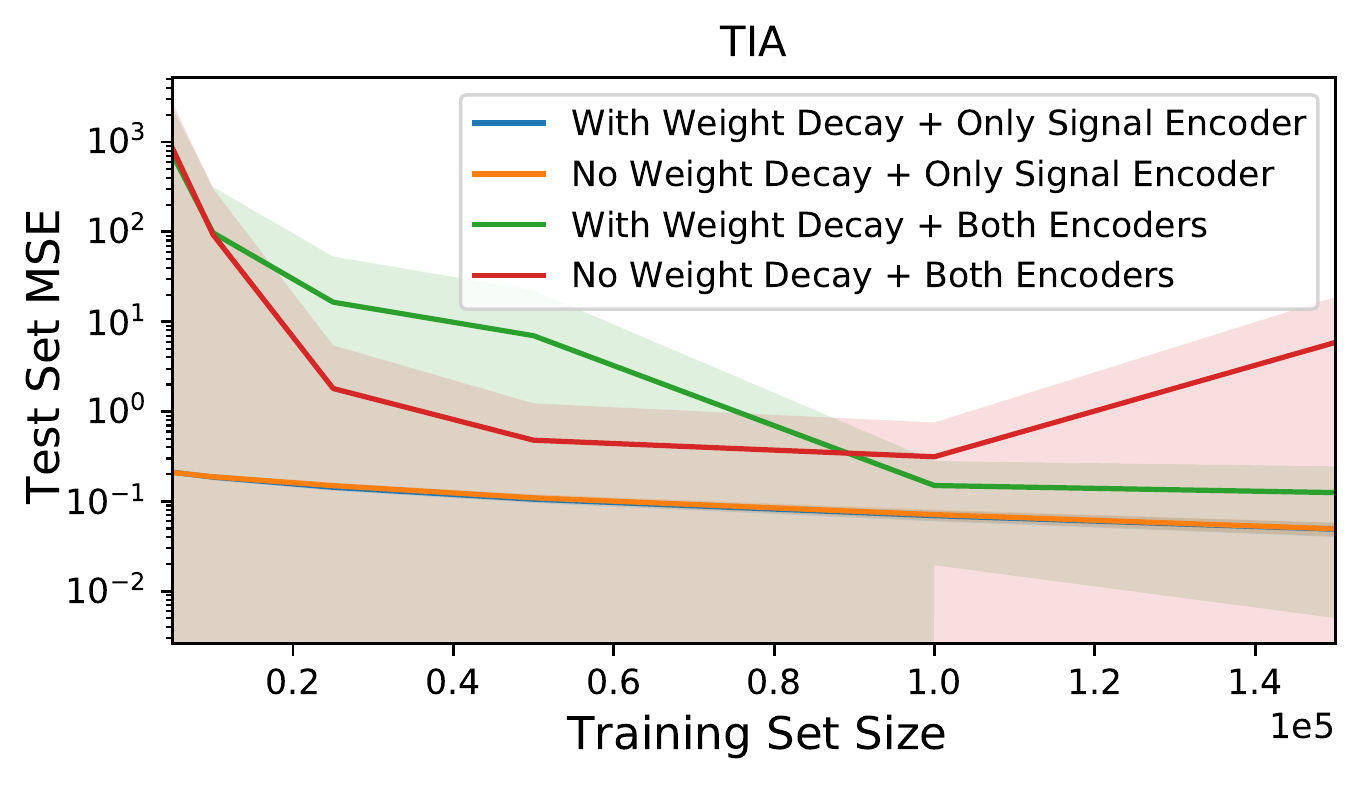}
    \caption{Performance of all TIA settings on \robodesk joint position regression. Only using the signal encoder is necessary for good performance.}
    \label{fig:robo-joint-tia}
    \end{minipage}
    \begin{minipage}[t]{\textwidth}
    \vspace{5pt}
    \centering
    \includegraphics[scale=0.34, trim=10 25 0 0 ]{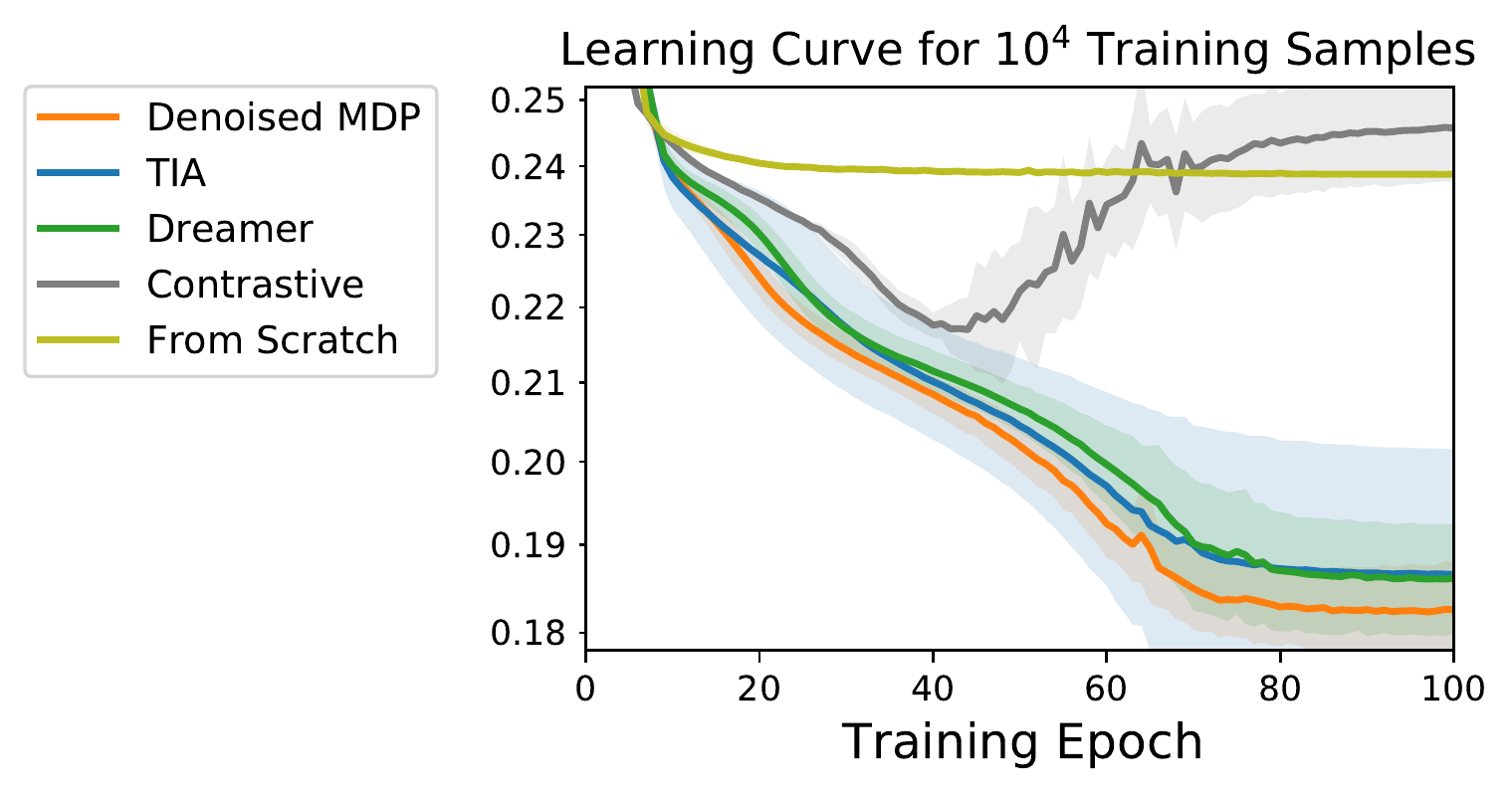}\hfill%
    \includegraphics[scale=0.34, trim=0 25 0 0 ]{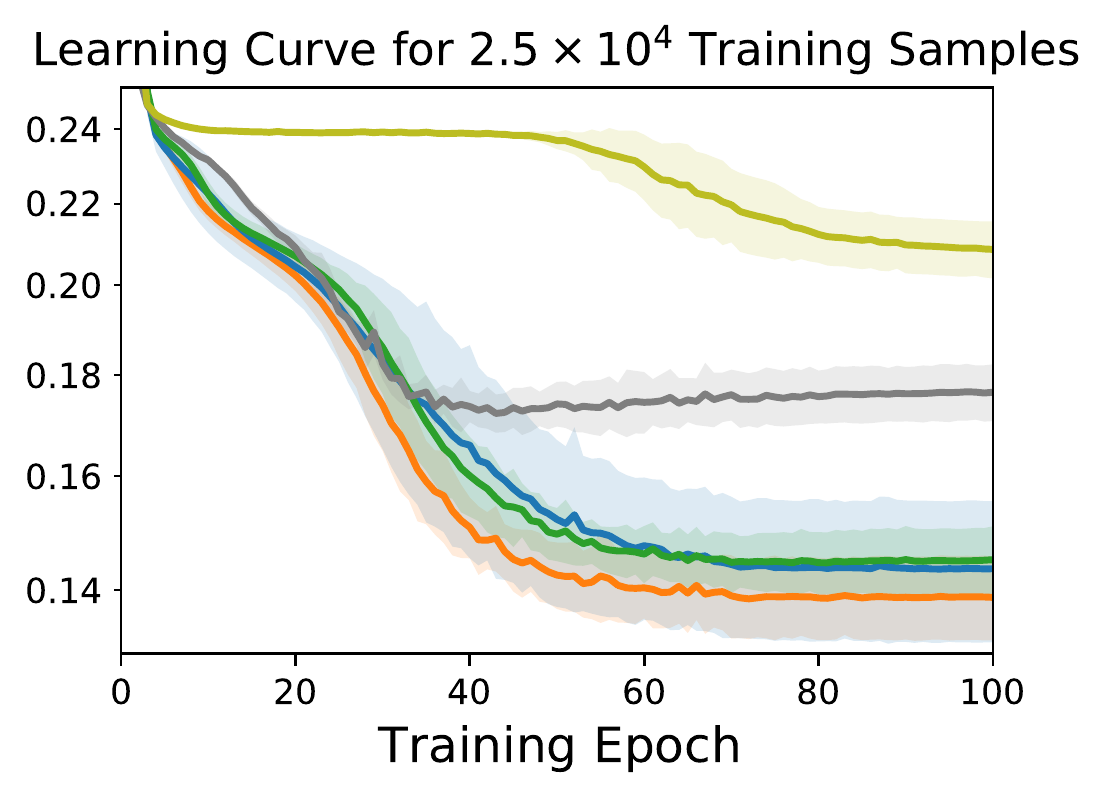}\hfill%
    \includegraphics[scale=0.34, trim=0 25 0 0 ]{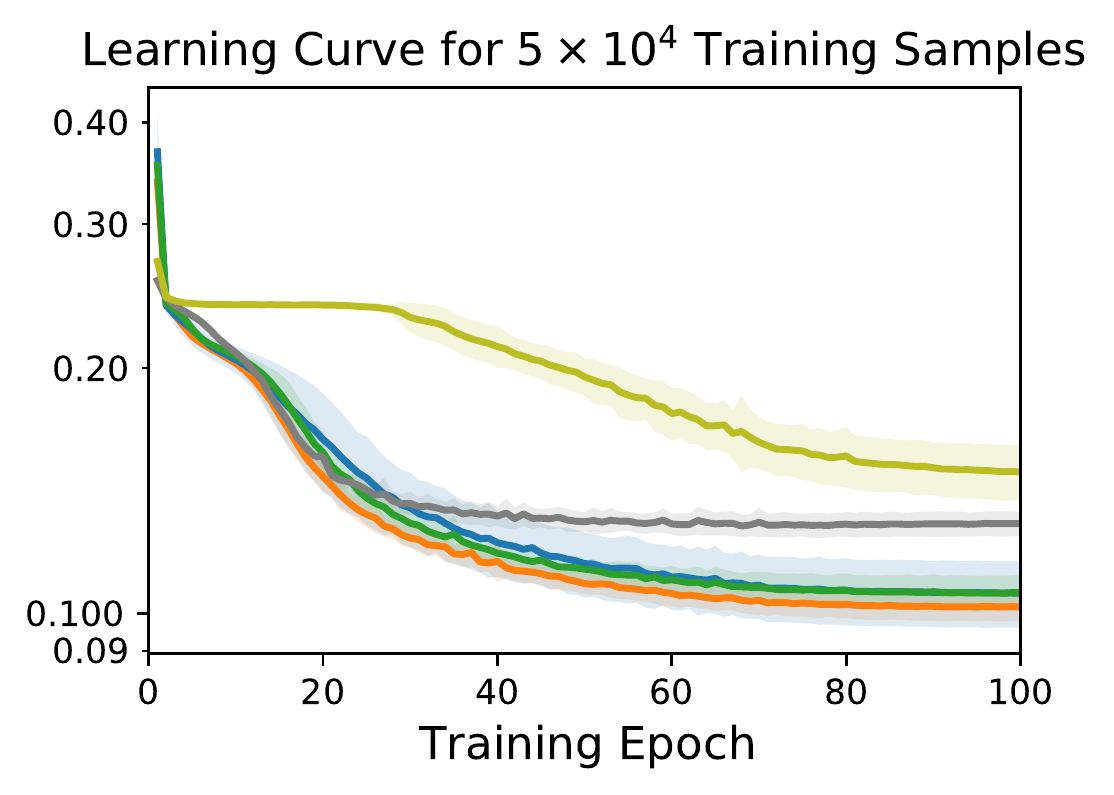}\hfill%
    \includegraphics[scale=0.34, trim=0 25 10 0 ]{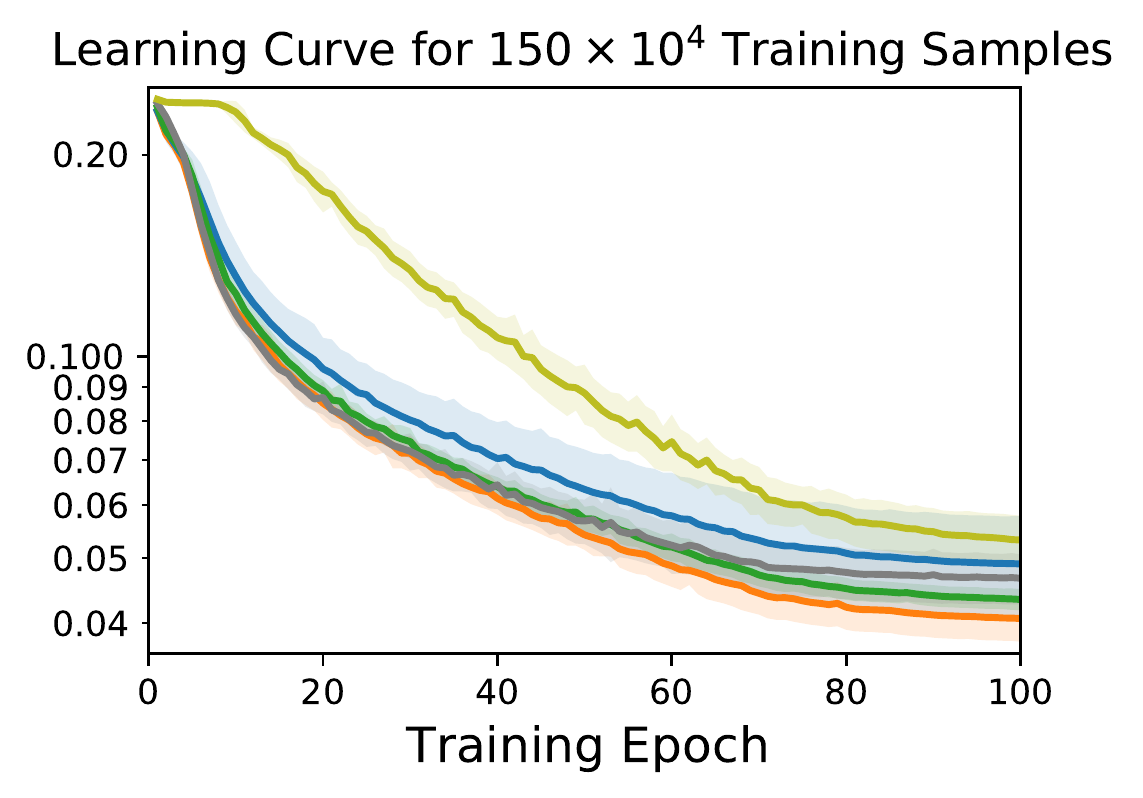}
    \caption{Training curve comparisons for the \robodesk joint position regression task across many training set sizes.}\label{fig:robo-joint-tr-curve-more}
    \end{minipage}
\end{figure}

\begin{figure}[t]
    \centering
    \begin{minipage}[t]{0.48\textwidth}
    \centering
    \includegraphics[width=0.9\linewidth, trim=0 20 0 0 ]{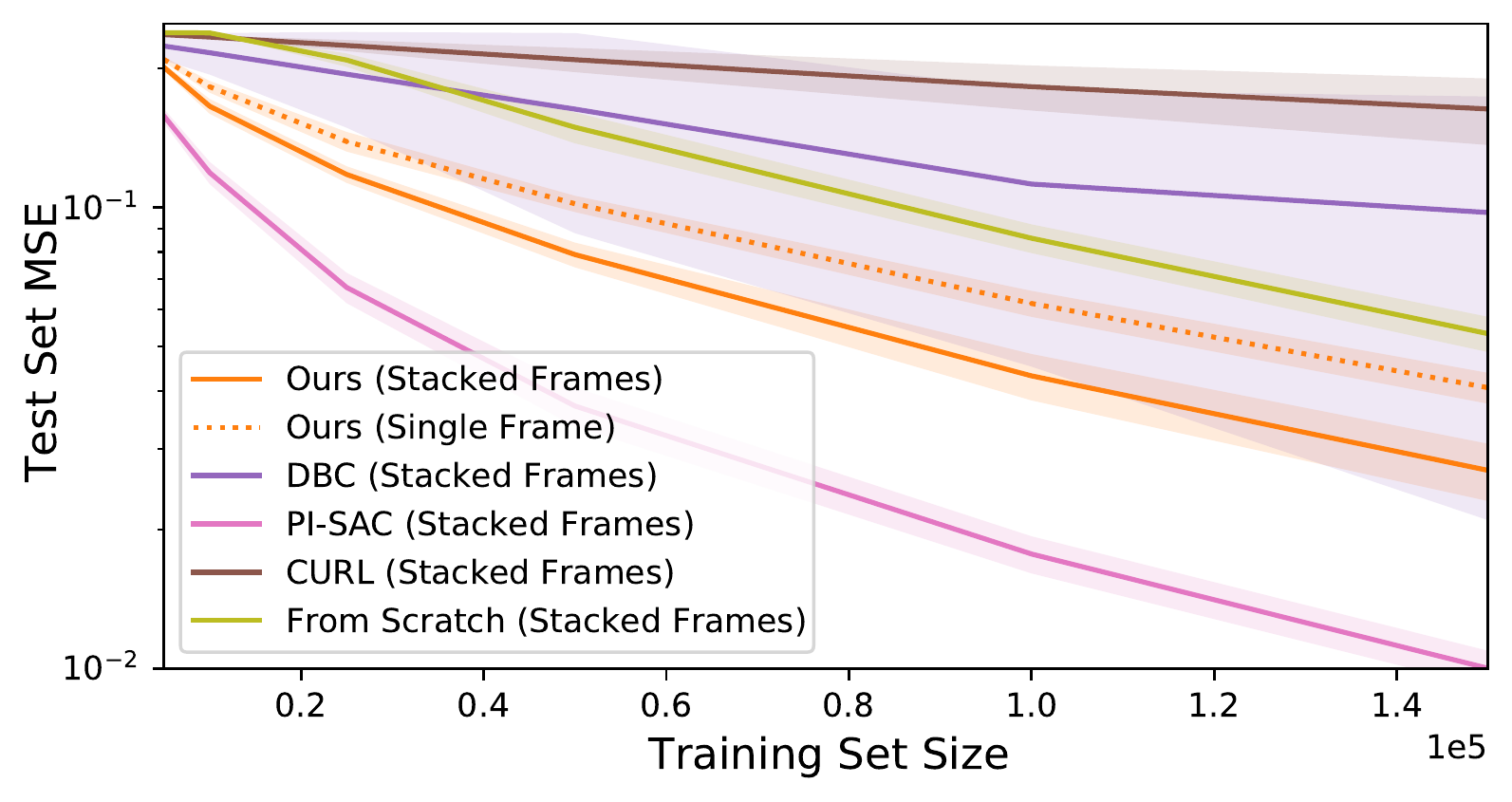}
    \caption{Performance comparison of finetuning from Denoised MDP encoders and frame-stacked encoders that take in $3$ consecutive frames. For Denoised MDP and training from scratch, the encoders \emph{take in only a single frame} and are applied for each of the frame, with output concatenated together before feeding to the prediction head.}\label{fig:robo-joint-framestack}
    \end{minipage}
    \hfill%
    \begin{minipage}[t]{0.48\textwidth}
    \centering
    \includegraphics[width=0.9\linewidth, trim=0 20 0 0]{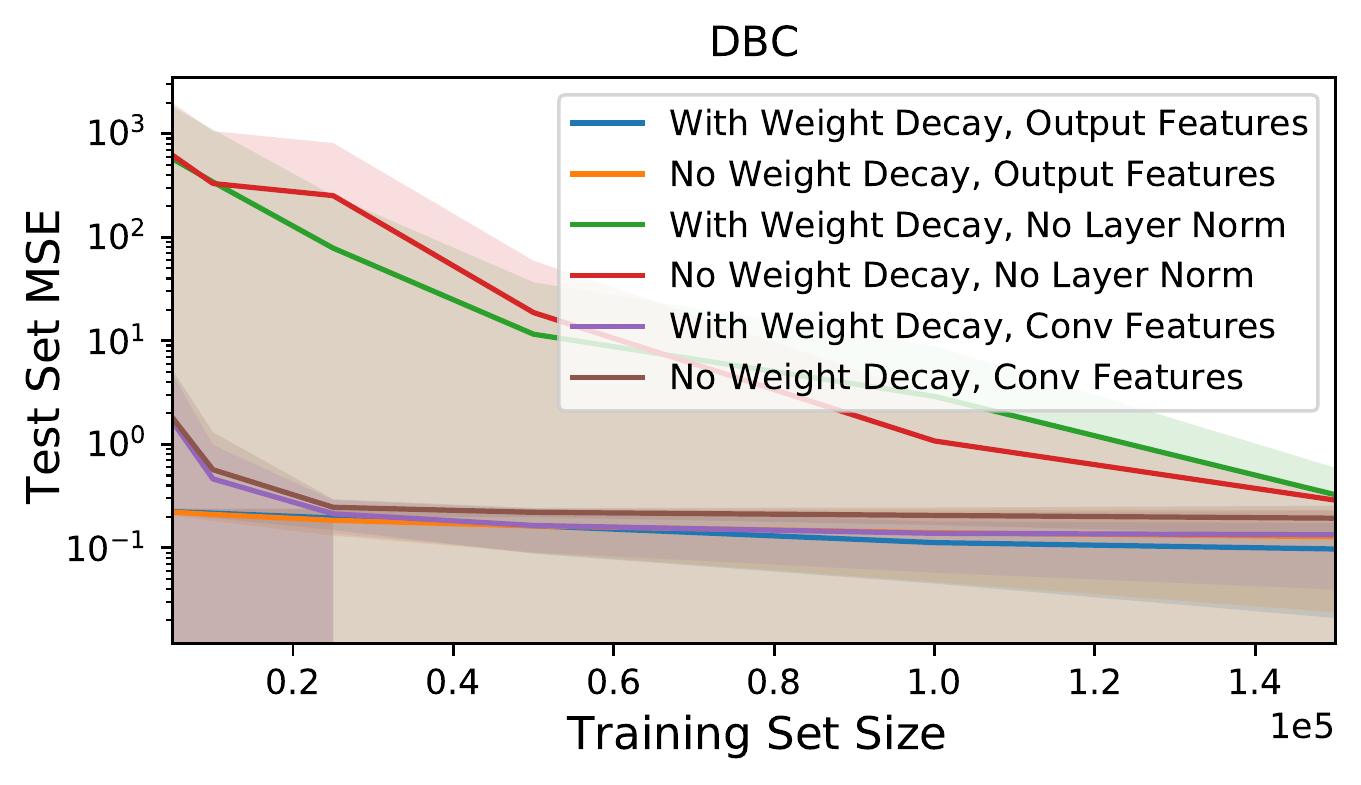}
    \caption{Performance of all DBC settings on \robodesk joint position regression. Using the output features (after layer normalization) is necessary for good performance.}
    \label{fig:robo-joint-dbc}
    \end{minipage}
    \begin{minipage}[t]{0.48\textwidth}
    \centering
    \includegraphics[width=0.9\linewidth, trim=0 20 0 0]{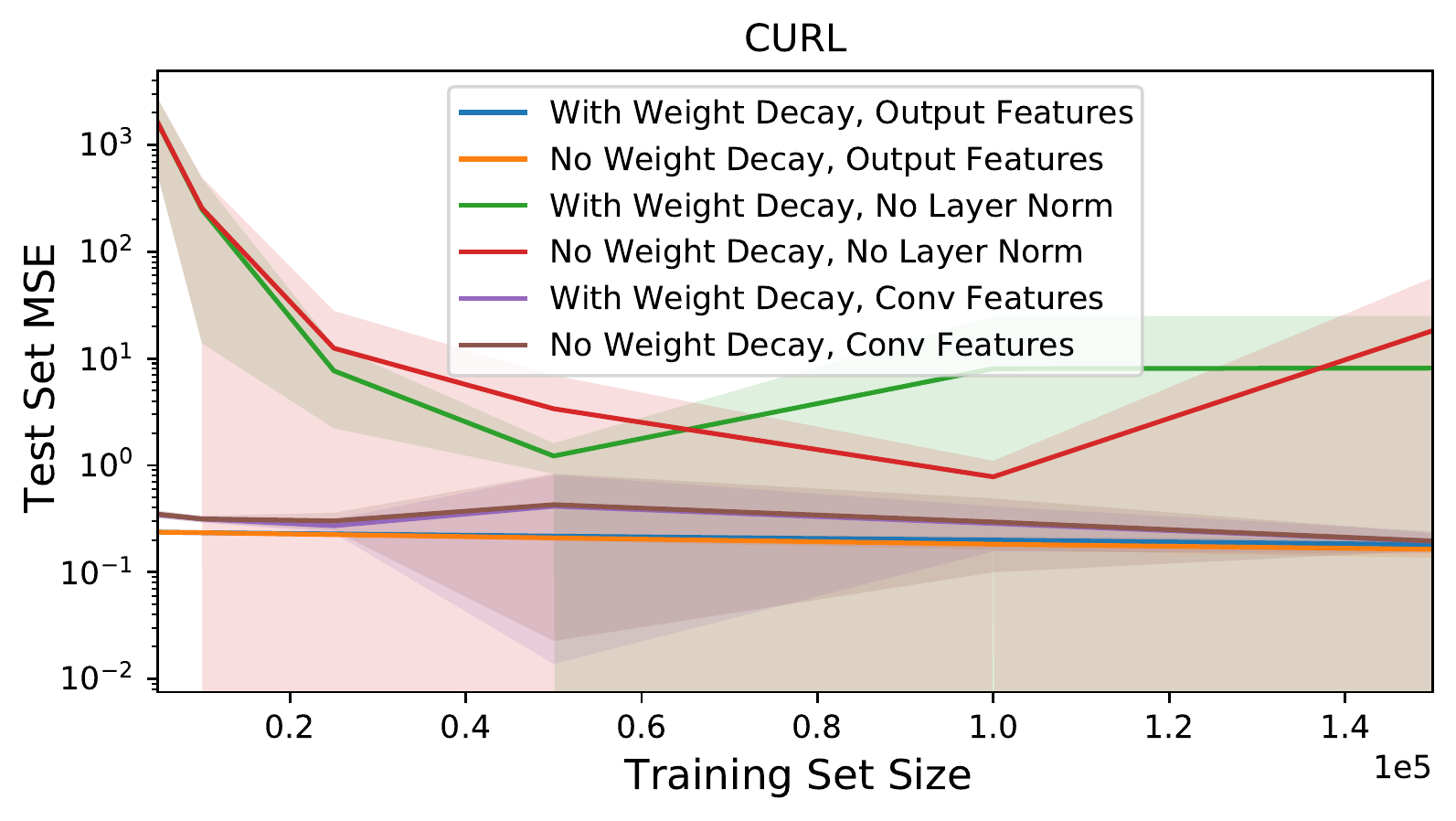}
    \caption{Performance of all CURL settings on \robodesk joint position regression. Using the output features (after layer normalization) is necessary for good performance.}
    \label{fig:robo-joint-curl}
    \end{minipage}\hfill%
    \begin{minipage}[t]{0.48\textwidth}
    \centering
    \includegraphics[width=0.9\linewidth, trim=0 20 0 0]{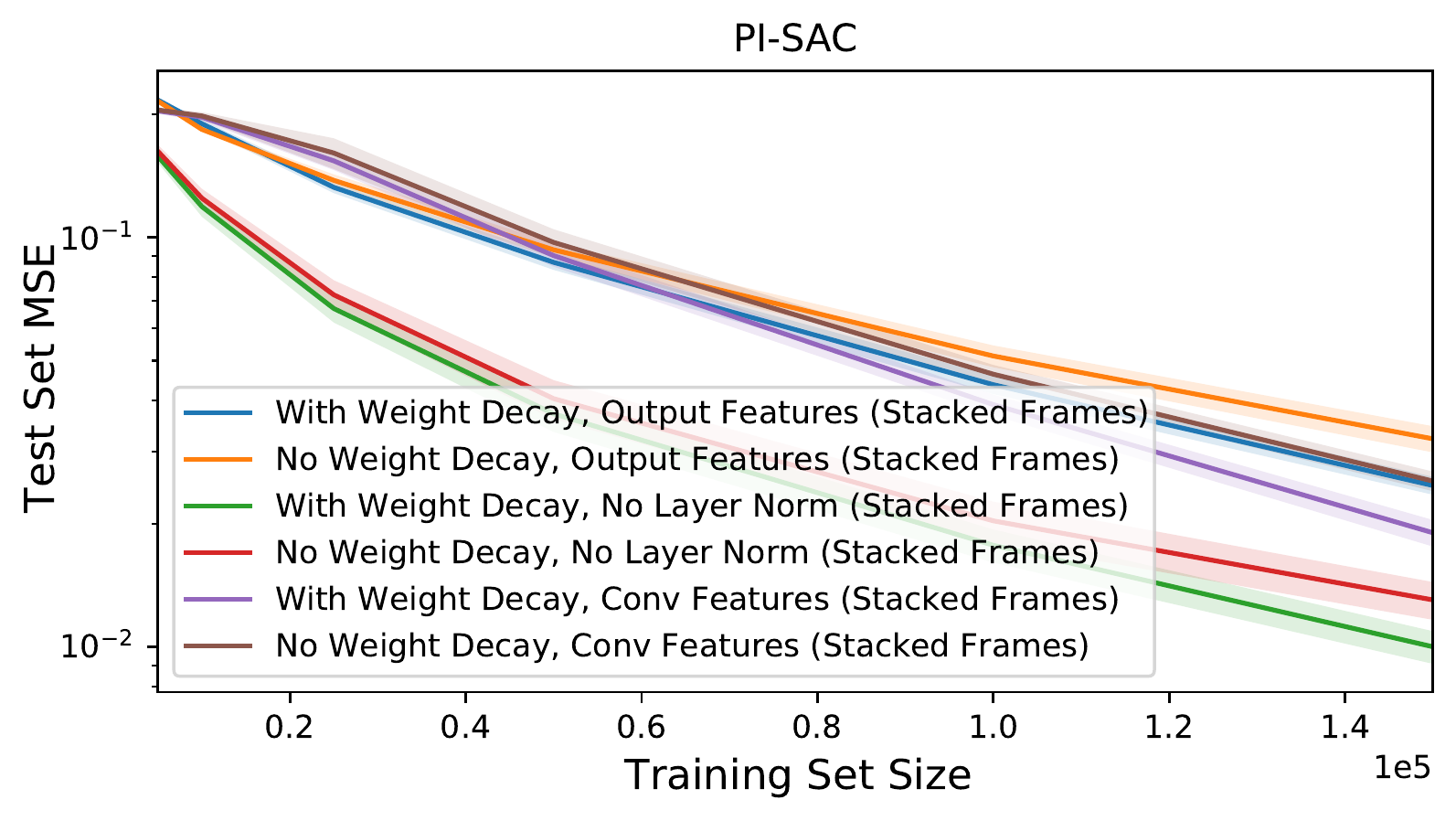}
    \caption{Performance of all PI-SAC settings on \robodesk joint position regression. Using the activations \emph{before layer normalization} gives best performance.}
    \label{fig:robo-joint-pisac}
    \end{minipage}
\end{figure}

\paragraph{Training details.} For this task, we jointly train the pre-trained backbone and a three-layer MLP head that has $256$ hidden units at each layer, with a learning rate of $8 \times 10^{-5}$. For finetuning from pretrained encoders, we follow common finetuning practice and apply a weight decay of $3\times 10^{-5}$ whenever it is helpful (all cases except CURL and training from scratch). See \Cref{fig:robo-joint-wd} for comparisons for weight decay options over all methods. \begin{itemize}
    \item For model-based RL, we take encoders trained with backpropagating via dynamics as the policy optimization algorithm.
    \item In training the contrastive encoder, for a (more) fair comparison with RL-trained encoders that are optimized over $10^6$ environment steps, we train contrastive encoders on $10^6$ samples, obtained in the exact same method of the training sets of this task.  In a sense, these contrastive encoders have the advantage of training on the exact same distribution, and seeing more samples (since RL-trained encoders use action repeat of $2$ and thus only ever see $0.5 \times 10^6$ samples).
    \item TIA has two sets of encoders. Using concatenated latents from both unfortunately hurts performance greatly (see \Cref{fig:robo-joint-tia}). So we use only the encoder for the signal latent.
\end{itemize}
We also compare training speeds over a wide range of training set sizes in \Cref{fig:robo-joint-tr-curve-more}. Denoised MDP encoders lead to faster and better training in all settings.

\new{\paragraph{Additional comparison with frame-stacking encoders.}
Other pretrained encoders (DBC, CURL and PI-SAC) take in stacked $3$ consecutive frames, and are not directly comparable with the other methods. To compare, we also try running Denoised MDP encoders on the $3$ consecutive frames, whose feature vector is concatenated before feeding into the head. The result in \Cref{fig:robo-joint-framestack} shows that our encoder outperforms all but PI-SAC encoders. Finally, for DBC, CURL and PI-SAC, we attempted evaluating intermediate features, features before the final layer normalization, and the output space, and find the last option best-performing for DBC and CURL, and the second option best-performing for PI-SAC (see \Cref{fig:robo-joint-dbc,fig:robo-joint-curl,fig:robo-joint-pisac}).  Therefore, we use these respective spaces, which arguably gives a further edge to these methods, as we essentially tune this additional option on test results.  Notably, these respective choices are often the only one achieving relatively good performance, highlighting the necessity of tuning for these methods.
}


\subsection{DeepMind Control Suite (DMC) Result Details}\label{sec:appendix-dmc-details}


\begin{figure*}[ht]
\centering
\includegraphics[scale=0.89, trim=25 10 90 10]{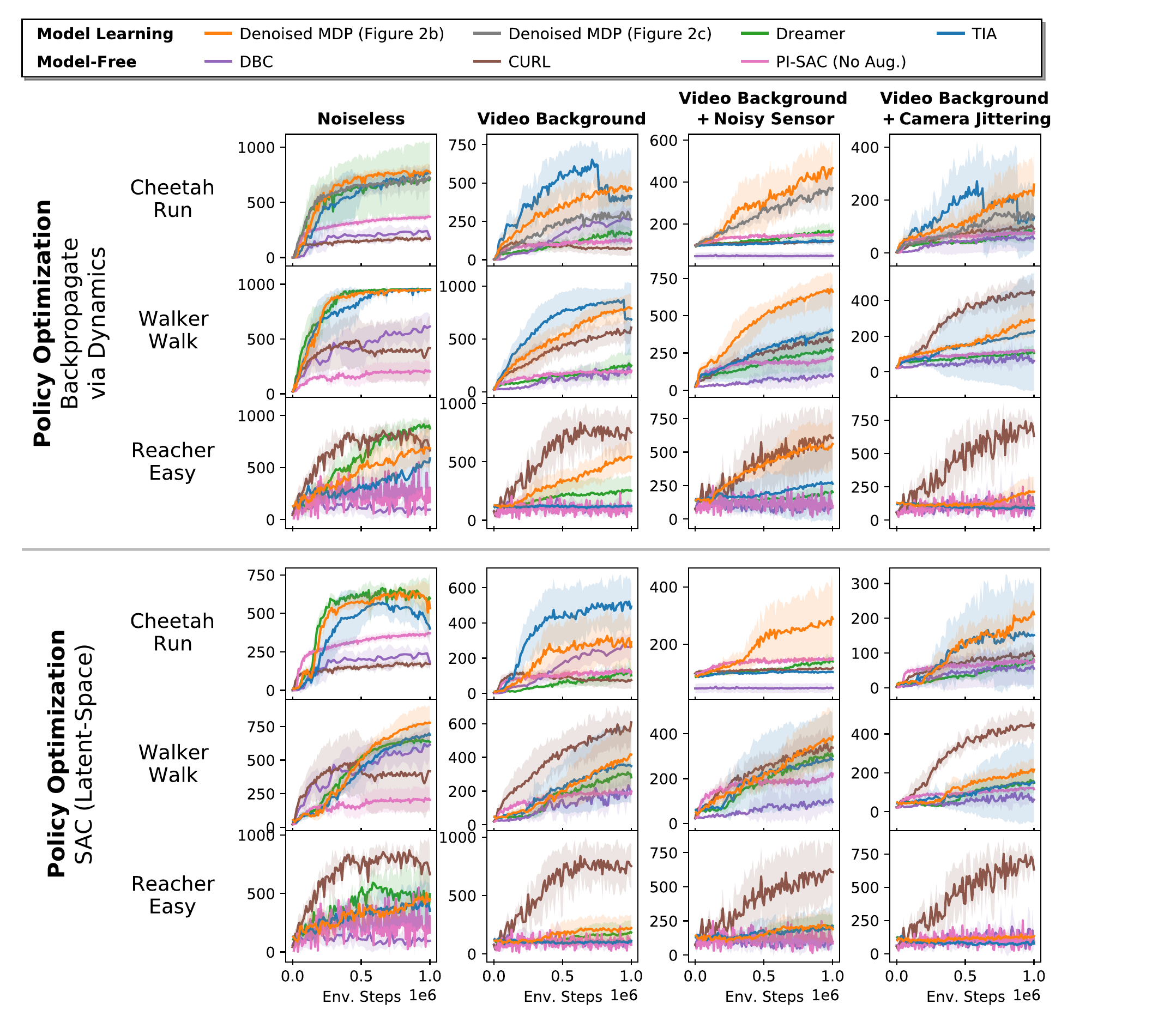}
\caption{Policy optimization results on DMC. Each plot focuses on a single task variant, showing total episode return versus environment steps taken. For three model-based approaches, we use two policy optimization choices to train on the learned model: \textbf{(top half)} backpropagate via learned dynamics and \textbf{(bottom half)} SAC on the learned MDP. We also compare with DBC, a model-free baseline. For an ``upper bound'' (not plotted due to presentation clarity), SAC on true state-space  (\ie,  optimal representation)  in $10^6$ environment steps reaches episode return $\approx 800$ on Cheetah Run variants,  $\approx 980$ on Walker Walk variants, and  $\approx 960$ on Reacher Easy variants. CURL's specific augmentation choice (random crop) potentially helps significantly for Reacher Easy (where the reacher and the target appear in random spatial locations) and \textbf{Camera Jittering}. However, unlike Denoised MDP, it does not generally perform well across all environments and noise variants. Top row shows a comparison between two Denoised MDP variants, one based on \Cref{fig:mdp-grid-xy} and another based on slighted modified \Cref{fig:mdp-grid-xyz}. See \Cref{sec:appendix-dmc-details} for details.  }\label{fig:dmc-rew}
\end{figure*}

\paragraph{Full policy optimization results.} \Cref{fig:dmc-rew} presents the full results on each DMC environment (task + variant). For environment, a comparison plot is made based on which policy learning algorithm is used with the model learning method (with model-free baselines duplicated in both). Such separation is aimed to highlight the performance difference caused by model structure (rather than policy learning algorithm). Across most noisy environments, Denoised MDP performs the best. It also achieves high return on noiseless environments.


\begin{figure}[ht]
\centering
\vspace{-6pt}\hspace*{-0.4em}%
\includegraphics[trim=22 23 10 0, clip, scale=0.7925]{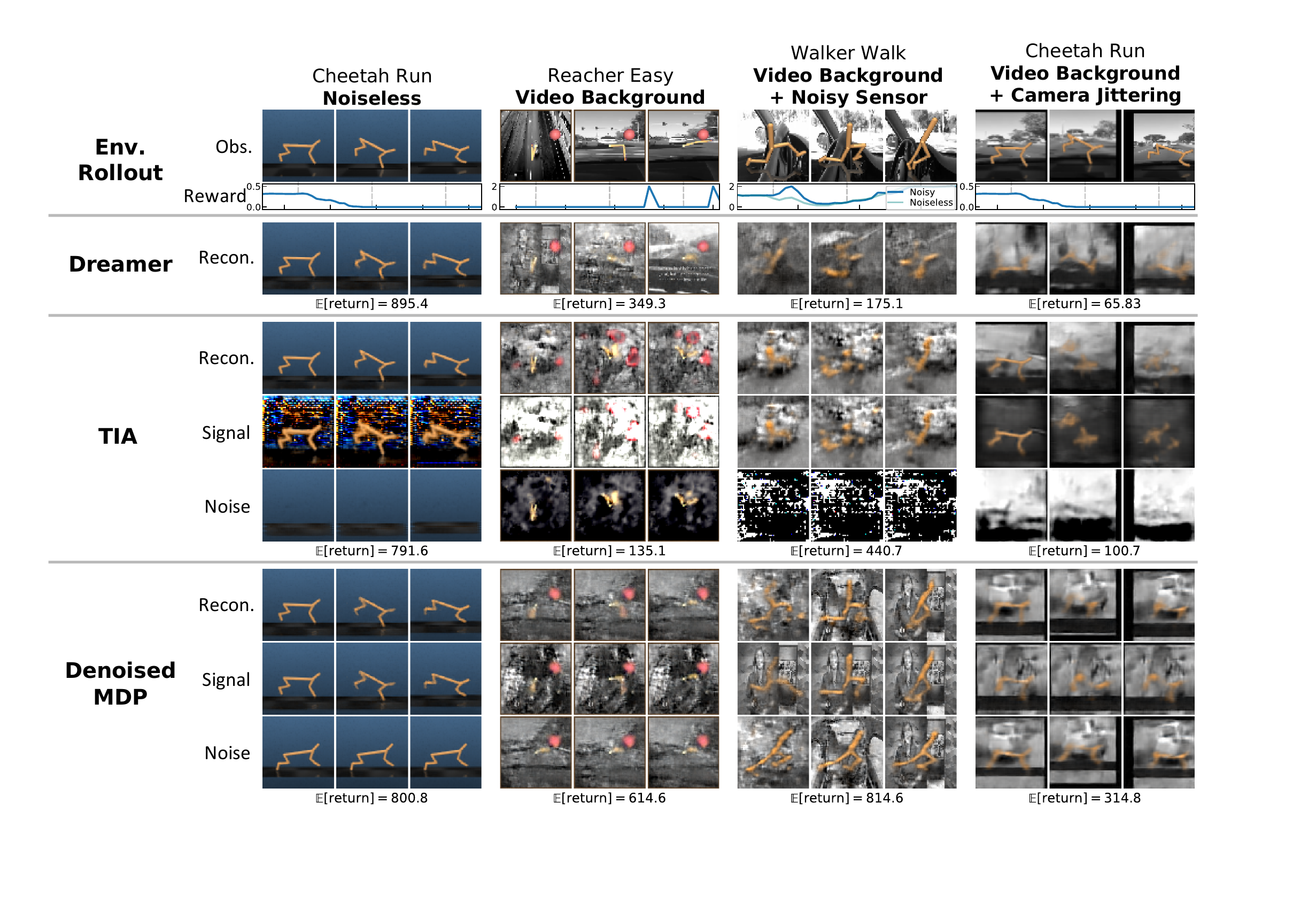}\vspace{-18pt}
\caption{Complete visualization of the different DMC variants  and factorizations learned by TIA and Denoised MDP. In addition to visualizations of \Cref{fig:dmc-recon}, we also visualize full reconstructions from Dreamer, TIA, and Denoised MDP.
}\label{fig:dmc-recon-full}
\end{figure}
\paragraph{Visualization of learned models.} \Cref{fig:dmc-recon-full} is the extended version of \Cref{fig:dmc-recon} in main text, with full reconstructions from all three models.  Please see the supplementary video for clearer visualizations.

\paragraph{Comparison between Denoised MDP variants.} We compare two Denoised MDP variants based on Cheetah Run environments with policy trained by packpropagating via learned dynamics. One variant is based on  \Cref{fig:mdp-grid-xy}, and is used for all our results in main paper. The other variant is based on \Cref{fig:mdp-grid-xyz} with a minor modification for $\cz$ prior (using $p^\theta(\czt \given \cxt, \czpt, a)$ rather than $p^\theta(\czt \given \cxt, \cyt, \czpt, a)$). Compared to the \Cref{fig:mdp-grid-xy} variant results, we run this new variant with the same $(120+20)$ latent size for $\cx$, but with $(70+10)$ latent size for both $\cy$ and $\cz$, to maintain a comparable number of parameters. All other hyperparameters are kept the same. This comparison is shown in the top row of \Cref{fig:dmc-rew}, where we see the \Cref{fig:mdp-grid-xy} variant often performing a bit better. We hypothesize that this may due to the more complex prior and posterior structure of \Cref{fig:mdp-grid-xyz}, which may not learn as efficiently. This also makes \Cref{fig:mdp-grid-xyz} variant needing longer (wall-clock) time to optimize, as mentioned above in \Cref{sec:compute-resource}.

\paragraph{TIA hyperparameters and instability.} We strictly follow recommendations of the original paper, and use their suggested value for each DMC task. \new{We also note that TIA runs sometimes collapse during training, leading to sharp drops in rewards. After closely inspecting the models before and after collapses, we note that in many cases, such collapses co-occur with sudden spikes in TIA's reward disassociation loss, which is implemented as an adversarial minimax loss, and the noise latent space instantly becomes degenerate (\ie, not used in reconstruction). We hypothesize that this adversarial nature can cause training instability. However, a few collapses do not co-occur with such loss spikes, which maybe alternatively due to that TIA model structure cannot model the respective noise types and that better fitting the model naturally means a degenerate noise latent space. }

\new{\paragraph{PI-SAC hyperparameters.} For each task, we use the hyperparameters detailed in the original paper \citep{lee2020predictive}. PI-SAC is usually run with augmentations. However, unlike CURL, augmentation is not an integral part of the PI-SAC algorithm and is completely optional. For a fair comparisons with other methods and to highlight the effect of the predictive information regularizer, the main mechanism proposed by PI-SAC, we do not use augmentations for PI-SAC. }


\begin{figure}[ht]
\begin{minipage}[t]{\linewidth}

\end{minipage}
\centering
\includegraphics[scale=0.89, trim=25 10 90 7]{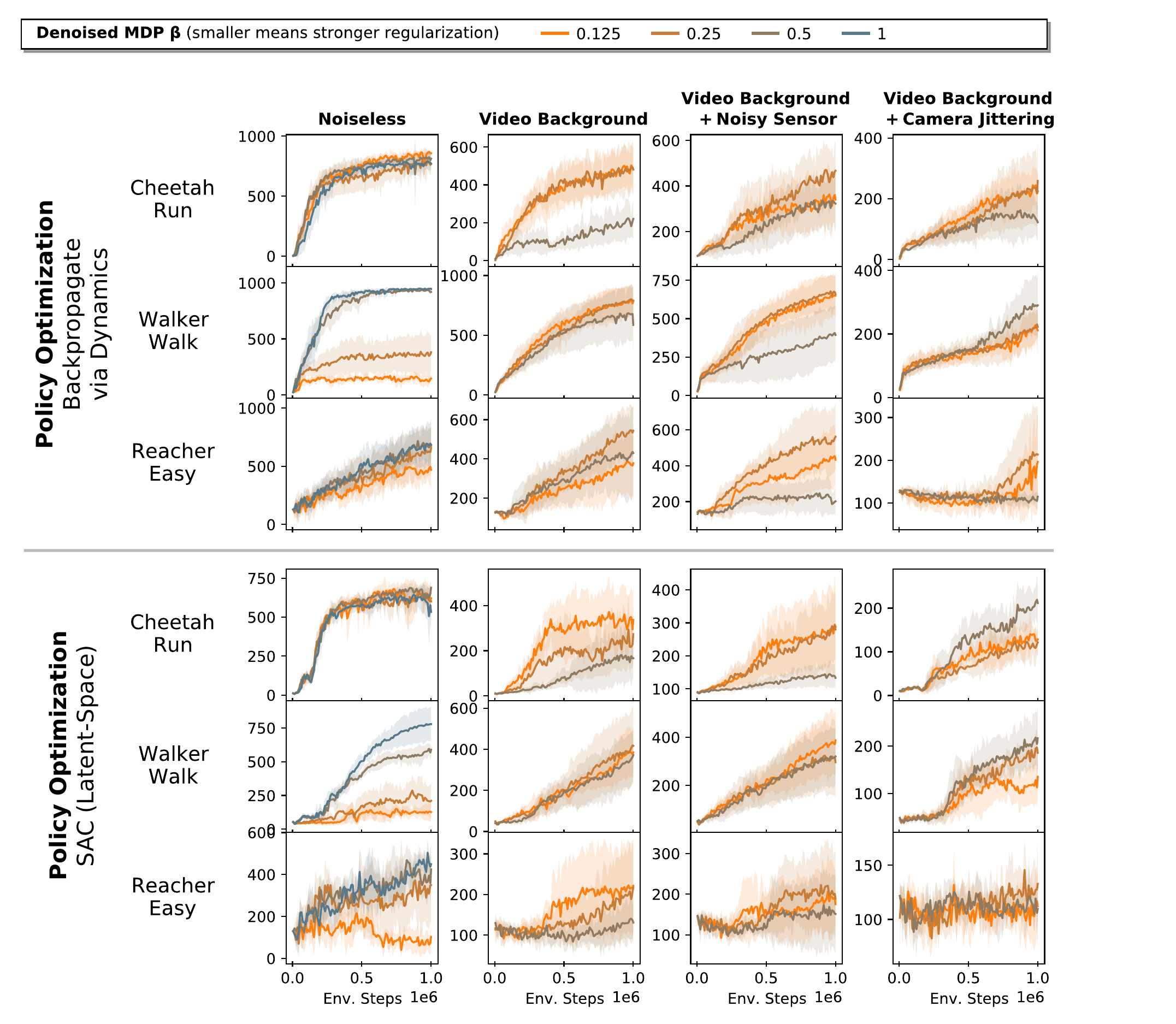}
\caption{Effect of choosing $\beta$ in Denoised MDP on DMC policy optimization results. Setting $\beta=1$ disables regularization and is \emph{only run on noiseless variants}.}\label{fig:dmc-rew-beta}
\begin{minipage}[t]{\linewidth}
\vspace{3pt}
\begingroup
\newcommand{\NA}{---}
\renewcommand{\arraystretch}{1.4}
\centering
\resizebox{
  0.89\width
}{!}{%
\small
\begin{tabular}{cccccc}
    \toprule 
    
    &
    & \multirow{2}{*}{\shortstack{\textbf{Noiseless}}}
    & \multirow{2}{*}{\shortstack{\textbf{Video Background}}}
    & \multirow{2}{*}{\shortstack{\textbf{Video Background}\\\textbf{+ Noisy Sensor}}}
    & \multirow{2}{*}{\shortstack{\textbf{Video Background}\\\textbf{+ Camera Jittering}}}
    \\
    & & & & & \\
    \midrule
    \multirow{3}{*}{\shortstack{\textbf{Policy Learning:}\\Backprop via Dynamics}}
    & Cheetah Run
    & 1
    & 0.125
    & 0.25 
    & 0.25
    \\
    & Walker Walk
    & 1
    & 0.25
    & 0.25
    & 0.5
    \\
    & Reacher Easy
    & 1
    & 0.25
    & 0.25
    & 0.25
    \\
    \midrule
    \multirow{3}{*}{\shortstack{\textbf{Policy Learning:}\\SAC (Latent-Space)}}
    & Cheetah Run
    & 1
    & 0.125
    & 0.125
    & 0.25 
    \\
    & Walker Walk
    & 1
    & 0.25
    & 0.125
    & 0.5
    \\
    & Reacher Easy
    & 1
    & 0.125
    & 0.25
    & 0.25
    \\
    \bottomrule
\end{tabular}%
}%
\vspace{-1.5pt}
\captionof{table}{$\beta$ choices for Denoised MDP results shown in \Cref{tbl:dmc-agg-rew,fig:dmc-rew}. We choose $\beta=1$ (\ie, disabling regularization) for all noiseless environments, and tuned others. However, as seen in \Cref{fig:dmc-rew-beta}, the results often are not too sensitive to small $\beta$ changes. }\label{table:dmc-beta}
\endgroup

\end{minipage}
\end{figure}

\paragraph{Denoised MDP hyperparameters.} For DMC, we always use fixed $\alpha=1$. $\beta$ can be tune according to amount of noises in environment, and to training stability. In \Cref{fig:dmc-rew-beta}, we compare effects of choosing different $\beta$'s. On noiseless environments, larger $\beta$ (\ie, less regularization) performs often better. Whereas on noisy environments, sometimes stronger regularization can boost performance. However, overall good performance can be obtained by usually several $\beta$ values. In \Cref{table:dmc-beta}, we summarize our $\beta$ choices for each environment in \Cref{table:dmc-beta}.

\end{document}